\documentclass{article}

\PassOptionsToPackage{numbers, compress}{natbib}
\usepackage{arxiv}

\usepackage[utf8]{inputenc} 
\usepackage[T1]{fontenc}    
\usepackage{hyperref}       
\usepackage{url}            
\usepackage{booktabs}       
\usepackage{amsfonts}       
\usepackage{nicefrac}       
\usepackage{microtype}      
\usepackage{xcolor}         

\usepackage{graphicx}
\usepackage{subcaption}
\usepackage{placeins}
\usepackage{tabularx}
\usepackage{caption}
\usepackage{tikz}
\usepackage{enumitem}

\usepackage{algorithm}
\usepackage[algo2e]{algorithm2e}

\usepackage{amsmath}
\usepackage{amssymb}
\usepackage{mathtools}
\usepackage{amsthm}
\usepackage{mathbbol}
\usepackage{bm}
\usepackage{upgreek} 
\usepackage{dsfont}
\usepackage{accents}

\usepackage[capitalize,noabbrev]{cleveref}
\theoremstyle{plain}
\newtheorem{theorem}{Theorem}[section]
\newtheorem{proposition}[theorem]{Proposition}

\theoremstyle{definition}
\newtheorem{definition}[theorem]{Definition}
\newtheorem{assumption}[theorem]{Assumption}

\theoremstyle{remark}

\def \E {\mathbb{E}}
\def \P {\mathbb{P}}
\definecolor{colorA}{HTML}{DADCFF} 
\definecolor{colorB}{HTML}{E2F0D9} 
\newcommand{\Enb}[1]{\mathbb{E}#1}
\newcommand{\norm}[1]{\|#1\|}
\newcommand{\abs}[1]{\left | #1 \right |}
\def \Var {\text{Var}}

\title{Trustworthy Selection of Heterogeneous Treatment Effect Estimators}

\author{Jiayi Guo$^{1}$,\, Zijun Gao$^{2}$\\
$^1$Peking University  \quad
$^2$Marshall School of Business, University of Southern California}

\begin{document}

\maketitle

\begin{abstract}
The rapid development of heterogeneous treatment effect (HTE) estimators has shifted practitioners' challenges from having no estimator to use, to navigating many competing options that are difficult to choose from.
We study the problem of selecting HTE estimator from a collection of candidates, where the key difficulty is that the individual treatment effect is fundamentally unobservable, rendering the task unsupervised (partially supervised).
We cast the estimator selection as a multiple testing problem and introduce a ground-truth-free, relative-error–based procedure that returns a set of estimators guaranteed to contain the winner with at least a prespecified probability. To make the output set as small as possible, we employ an exponentially weighted scheme that adaptively prioritizes the most competitive candidates. We further design a two-way sample-splitting strategy that decouples the nuisance estimation in the relative error construction and the learning of the exponential weight, which enables a stability-based central limit theorem and further yields the asymptotic error control. Empirically, our method provides reliable error control while reducing false selections relative to commonly used heuristics across the ACIC 2016, IHDP, and Twins benchmarks.
\end{abstract}

\section{Introduction}
The estimation of heterogeneous treatment effects (HTEs) has become a central topic across statistics, econometrics, and machine learning, with applications ranging from personalized medicine to policy evaluation~\cite{imbens2015causal, wager2018estimation, Hernan-Robins2020}. A growing body of work has proposed flexible estimators to capture individual-level treatment heterogeneity, including tree-based methods~\cite{wager2018estimation}, representation-learning approaches~\cite{shalit2017estimating, hassanpour2019counterfactual}, and meta-learners~\cite{Kunzel-etal2019, nie2021quasi}.
Despite this abundance of methods, determining which estimator performs best for a given application remains an open and underexplored problem~\cite{gao2021, guo2026relative}. A reliable selection mechanism is crucial for practitioners~\cite{curth2023searchinsightsmagicbullets}, as choosing suboptimal estimators can directly affect downstream decision-making~\cite{frauen2025treatment}.

Evaluating or comparing HTE estimators is inherently difficult because the ground truth is unobservable: for each individual, only one potential outcome is realized~\cite{holland1986statistics}, while the HTE is defined as the difference between two. 
Due to the fundamental unobservability of the treatment effect, comparing two HTE estimators is already challenging, and the difficulty is further exacerbated when a collection of estimators are being compared simultaneously. 
To our knowledge, most papers that compare multiple HTE estimators rely on simulated ground-truth and use them to compute metrics such as the Precision in Estimation of Heterogeneous Effect (PEHE) and Average Treatment Effect (ATE)~\cite{crabbe2022benchmarking, neal2021realcauserealisticcausalinference}. 
However, these evaluation metrics are subject to fundamental limitations: ground-truth are unavailable in real-world observational studies, and simulated values depend critically on the chosen data-generating process and offer no formal statistical guarantees.

In this paper, we develop a method for selecting the best heterogeneous treatment effect estimator that operates without ground-truth and provides provable error control.
Explicitly, we view the task of estimator selection as the statistical inference problem of the argmin of an unobserved loss, where the goal is returning a set of estimators guaranteed to include the best one.
To avoid relying on ground truth, we embed the relative-error-based pairwise comparison method of Gao~\citep{Gao2024} within the multiple-testing framework.
To increase the accuracy of the selected winner, i.e., to decrease the number of selected estimators, we construct an adaptive exponentially weighted statistic that aggregates relative errors across candidates motivated by Zhang et al.~\citep{zhang2024winners, kim2025locallyminimaxoptimalconfidence}.
Since the weight learning in the argmin inference and the nuisance estimation in the relative error construction can compromise each other's validity, we design a two-way sample-splitting scheme that separates these two components.
Theoretically, we establish the asymptotic validity of our method via a stability-based argument applied to the proposed two-way sample splitting.

The main contributions of this work are summarized as follows:
\begin{itemize}
    \item We formulate HTE estimator selection as a multiple-testing problem that enables uncertainty quantification.
    \item We develop a ground-truth-free testing procedure that, through a tailored adaptive weighting combined with two-way sample splitting, accurately selects the best-performing estimator. 
    \item We theoretically establish the asymptotic error control of our proposal based on a second-order stability argument.
    \item We conduct extensive experiments on benchmark datasets to demonstrate the effectiveness of the proposed method against commonly-used baselines. 
\end{itemize}

\textbf{Organization.} In Section 2, we introduce the causal framework, the relative error metric, and the formalization of the estimator selection problem. Section 3 details our proposed cross-fitted exponentially weighted testing procedure and establishes its theoretical guarantees for FWER control. Section 4 presents extensive experimental results on benchmark datasets. Finally, Section 5 concludes the paper. Due to space limitations, the review of related work, along with detailed proofs and additional experimental settings, is provided in the Appendix.
\vspace{-3pt}
\section{Preliminaries}
\vspace{-2pt}
\subsection{Causal Background}
We begin by introducing the notations and framework for our study.  
For each individual $i$, let $W_i \in \{0, 1\}$ denote the binary treatment indicator, where $W_i=1$ represents receiving the treatment and $W_i=0$ denotes control.  
Let $X_i \in \mathbb{R}^d$ be the vector of pre-treatment covariates.
Under the potential outcomes framework~\cite{Rubin1974, Neyman1990}, we denote by $Y_i(1)$ and $Y_i(0)$ the potential outcomes corresponding to $W_i=1$ and $W_i=0$, respectively.  
We observe the outcome
\(
Y_i = W_i Y_i(1) + (1-W_i)Y_i(0).
\)

The individual treatment effect (ITE) is defined as $Y_i(1) - Y_i(0)$, representing the treatment effect for individual $i$.  
Because only one potential outcome is observed for each unit, ITEs are not identifiable without additional assumptions~\cite{Hernan-Robins2020}.  
A common surrogate is the \emph{conditional average treatment effect (CATE)}, defined as
\(
\tau(x) = \E[Y_i(1) - Y_i(0) \mid X_i = x],
\)
which characterizes how the treatment effect varies across individuals with different covariate profiles.

\begin{assumption}[\cite{rosenbaum1983central}]
\label{assump1}
(i) $(Y_i(0), Y_i(1)) \perp\!\!\!\perp W_i \mid X_i$; \quad  
(ii) $0 < e(x) \triangleq \P(W_i=1 \mid X_i=x) < 1$ for all $x \in \mathcal{X}$,  
where $e(x)$ denotes the propensity score.
\end{assumption}


Under Assumption~\ref{assump1}, the CATE is identified as
$\tau(x) = \mu_1(x) - \mu_0(x)$,
where $\mu_a(x) = \E[Y_i \mid W_i=a, X_i=x]$ for $a\in\{0,1\}$.  
As mentioned in the introduction, a rich literature has proposed various estimators of $\tau(x)$, such as causal forests~\cite{wager2018estimation} and representation learning approaches~\cite{shalit2017estimating}.

\subsection{Evaluation Metrics: Relative Error}

We next focus on evaluating a collection of heterogeneous treatment effect estimators. 
For any two estimators $\hat{\tau}_1$ and $\hat{\tau}_2$, their difference in performance can be quantified by
\begin{align*}
\delta(\hat{\tau}_1, \hat{\tau}_2)
&\triangleq \E[(\hat \tau_1(X) - \tau(X))^2] - \E[(\hat \tau_2(X) - \tau(X))^2]\\
& = \E\!\left[\hat{\tau}_1^2(X) - \hat{\tau}_2^2(X)
- 2(\hat{\tau}_1(X) - \hat{\tau}_2(X))\,\tau(X)\right].
\end{align*}
Following Gao~\citep{Gao2024}, we refer to $\delta(\hat{\tau}_1, \hat{\tau}_2)$ as the \emph{relative error}. 
While estimating an estimator’s absolute error $\phi(\hat\tau) = \E[(\hat\tau(X)-\tau(X))^2]$ is common, 
it has been shown that relative error enjoys stronger theoretical justification and improved empirical stability. 
Intuitively, relative error depends only on the first-order term of the unobserved $\tau(X)$, thereby reducing the influence of its estimation error. 

Given observed data $\{Z_i = (X_i, Y_i, W_i)\}_{i=1}^n$, we estimate the relative error through a one-step correction estimator:
\begin{align}
& \hat \delta (\hat \tau_1, \hat \tau_2)
\triangleq \frac{1}{n} \sum_{i=1}^n \hat t (Z_i; \hat \tau_1, \hat \tau_2) = \frac{1}{n} \sum_{i=1}^n 
\big[\hat{\tau}_1^2(X_i) - \hat{\tau}_2^2(X_i)\big] - 2 \big( \hat{\tau}_1(X_i) - \hat{\tau}_2(X_i) \big) \notag\\
& \cdot \left(
\frac{W_i (Y_i - \tilde{\mu}_1(X_i))}{\tilde{e}(X_i)} + \tilde{\mu}_1(X_i) - \frac{(1 - W_i)(Y_i - \tilde{\mu}_0(X_i))}{1 - \tilde{e}(X_i)}
- \tilde{\mu}_0(X_i)
\right) \label{relative_def}
\end{align}
where $\tilde{\mu}_1(X)$, $\tilde{\mu}_0(X)$ denote estimated outcome regressions and $\tilde{e}(X)$ denotes the estimated propensity score, all obtained from the validation dataset.
To ensure robustness and avoid overfitting, we employ a $K$-fold cross-fitting scheme separating nuisance estimation and relative error computation:
the sample is partitioned into folds $\{D_1,\ldots,D_K\}$; 
nuisance estimators \(\hat \eta_{D_{-k}}:=\{\tilde{\mu}_{0,D_{-k}},\tilde{\mu}_{1,D_{-k}},\tilde{e}_{D_{-k}}\}\) are trained on $D_{-k}$ when computing relative error on $D_k$. 
The final $\hat{\delta}(\hat{\tau}_1,\hat{\tau}_2)$ is obtained by averaging across folds:
\[
\hat{\delta}(\hat{\tau}_1,\hat{\tau}_2)
=\frac{1}{K}\sum_{k=1}^K
\frac{1}{|D_k|}\sum_{Z_i\in D_k}
\hat{t}(Z_i; \hat{\tau}_1,\hat{\tau}_2; \hat \eta_{D_{-k}}).
\]
Throughout this work, we choose \(K=2\) for convenience, which is parallel to \(K > 2\).

{
Gao~\citep{Gao2024} establishes the asymptotic normality of the estimated relative error \(\hat \delta(\hat \tau_1, \hat \tau_2)\).
\begin{equation} \label{delta_clt}
\frac{\hat \delta(\hat \tau_1, \hat \tau_2) - \delta(\hat \tau_1, \hat \tau_2)}{\sqrt{\hat V \delta(\hat \tau_1, \hat \tau_2)) / n}} \xrightarrow{d} \mathcal{N}(0, 1)
\end{equation}
where \(\hat V( \delta(\hat \tau_1, \hat \tau_2))\) is the empirical variance of \(\hat  \delta( \tau_1, \hat \tau_2)\),
as long as the following assumptions are satisfied.
}
\begin{assumption}[Bounded outcomes and overlap] \label{a1}
$Y$ is bounded, and there exists $\eta > 0$ such that
\[
\eta < e(X) < 1 - \eta .
\]
\end{assumption}

\begin{assumption}[Consistency of nuisance estimators] \label{a2}
The nuisance function estimators obtained from the test data satisfy
\[
\|\tilde{\mu}_a(X) - \mu_a(X)\|_2, \ \|\tilde{e}(X) - e(X)\|_2 = o_p(1),
\]
\[
\mathbb{E}\!\big[ (\tilde{\mu}_a(X) - \mu_a(X))(\tilde{e}(X) - e(X)) \big] = o_p(n^{-1/2})
\]
for \(a = 0, 1\).
\end{assumption}

\begin{assumption}[Non-degenerate relative error]
\label{a3}
The two estimators are not identical almost surely:
\[
\mathbb{E}\!\big[(\hat{\tau}_1(X) - \hat{\tau}_2(X))^2\big] \neq 0,
\]
and $\hat{\tau}_1(X) - \hat{\tau}_2(X)$ is not almost surely constant.
\end{assumption}

{For our theoretical analysis, we additionally impose a mild stability condition on the nuisance estimators under sample splitting.
}

\begin{assumption}[Stability of nuisance estimators]\label{a4}
The prediction performance of the nuisance estimators (especially the conditional outcomes) is stable.
\[
\|\tilde{\mu}_a (X ; S_{tr}) - \tilde{\mu}_a (X ; S_{tr}^{(l)}) \| = O(n^{-1}), \; \text{as} \; n \to \infty
\]
\begin{align*}
&\|\tilde{\mu}_a (X ; S_{tr}) - \tilde{\mu}_a (X ; S_{tr}^{(r)}) - \tilde \mu_a(X; S_{tr}^{(s)}) + \tilde \mu_a(X; S_{tr}^{(s,r)}) \| = o(n^{-1}), \; \text{as} \; n \to \infty \notag
\end{align*}
where \(S_{tr}^{(l)}\) is the same as the training dataset except that the $l$-th observation is replaced by an i.i.d. copy.
\end{assumption}

\subsection{Problem Formalization}

Building on the previous discussion of evaluation metrics, we now formalize the problem of comparing multiple CATE estimators.  
Suppose we have a collection of $p$ candidate estimators, each trained on an independent training sample:
\(
\{\hat{\tau}_1(x), \hat{\tau}_2(x), \dots, \hat{\tau}_p(x)\}.
\)
Given a test dataset $\{(X_i, W_i, Y_i)\}_{i=1}^n$ drawn i.i.d.\ from an underlying super-population $\mathbb{P}$, our goal is to identify the estimator that achieves the highest accuracy on the test sample.

Formally, let $\mathcal{S} = \{\hat{\tau}_{(1)}\}$ denote the set containing the true best-performing estimator $\hat{\tau}_{(1)}$.  
We aim to construct a data-driven selection set $\hat{\mathcal{S}}$ such that
\begin{align}
\lim_{n \to \infty} \P\big(\hat{\tau}_{(1)} \in \hat{\mathcal{S}}\big) = \alpha, \label{eq:goal}
\end{align}
where $\alpha$ is the prescribed significance level.  

Under Assumption~\ref{a3}, the best-performing estimator is unique, allowing us to reduce the selection problem to a series of marginal hypothesis tests.  
Specifically, for each estimator $r = 1, \dots, p$, we test whether it dominates all others in terms of relative error:
\begin{align}
H_0^r : \delta(\hat{\tau}_r, \hat{\tau}_s) < 0, \quad \forall s \neq r, \quad H_1^r : \exists\, s \neq r \text{ such that } \delta(\hat{\tau}_r, \hat{\tau}_s) > 0. \notag
\end{align}
Controlling the familywise error rate (FWER), i.e., the probability of making at least one Type I error across these $p$ hypotheses ensures that condition~\eqref{eq:goal} holds asymptotically.

For completeness, a summary of the notations used in this section is provided in Appendix~\ref{app-notation}.

\section{Testing Procedures}
{
Before introducing our proposed method, we first describe a simple and intuitive naive approach. We begin by introducing some notation. For a fixed $r\in\{1,\dots,p\}$, define the index set $\mathcal{I}_r \coloneqq \{1,\dots,p\}\setminus\{r\}$ and the vector of true and estimated pairwise relative errors
\[\delta_r
\triangleq
\big(\,\delta(\hat{\tau}_r,\hat{\tau}_s)\,:\, s\in\mathcal{I}_r\big)^\top,
\]
\[
\hat{\delta}_r
\triangleq
\big(\,\hat{\delta}(\hat{\tau}_r,\hat{\tau}_s)\,:\, s\in\mathcal{I}_r\big)^\top
\ \in \ \mathbb{R}^{K-1}.\]
Let $\Sigma_r \in \mathbb{R}^{(K-1)\times(K-1)}$ denote the covariance matrix of $\hat{\delta}_r$ and $\hat{\Sigma}_r$ its empirical estimator. 
}
\subsection{A Naive Approach} \label{sec-naive}
\noindent \textbf{Testing rule.} Leveraging the asymptotic normality, we can naturally construct a simple max–statistic test to determine whether a given estimator $\hat{\tau}_r$ is the true winner among all candidates.  Intuitively, if $\hat{\tau}_r$ is truly the winner, then all components of $\delta_r$ are nonpositive, 
and the test statistics
\[
S_{r,s} \triangleq \frac{\hat{\delta}(\hat{\tau}_r,\hat{\tau}_s)}{\hat{\sigma}_{r,s}}, \; s\in\mathcal{I}_r,
\quad\text{and}\quad
S_r^{\max} \triangleq \max_{s\in\mathcal{I}_r} S_{r,s},
\]
where $\hat{\sigma}_{r,s}^2 \triangleq (\hat{\Sigma}_r)_{ss}$, should rarely exceed its null quantile.  Hence, we test whether $S_r^{\max}$ is large enough to suggest that some $\delta(\hat{\tau}_r, \hat{\tau}_s) > 0$.

By multivariate CLT, we have 
\((S_{r,1}, \dots, S_{r,K-1})^\top \;\dot{\sim}\; \mathcal{N}(\delta_r, \Sigma_r).\)
{Because $\Sigma_r$ is generally non-diagonal, the null distribution of $S_r^{\max}$ has no closed form.
We therefore obtain the critical value $c_{1-\alpha}^{(r)}$ via a parametric bootstrap based on the estimated covariance $\hat{\Sigma}_r$.
We reject $H_0^{(r)}$ if $S_r^{\max} > c_{1-\alpha}^{(r)}$.  Accordingly, we define the selection set of the naive method as $\hat{\mathcal{S}}_{naive} \triangleq \{r \in [K] : S_r^{max} \le c_{1-\alpha}^{(r)}\}$.
The complete procedure is summarized in Algorithm~\ref{alg:naive} in the appendix.} 

\noindent \textbf{FWER control.} The familywise error rate (FWER) is automatically controlled at level $\alpha$, since each null $H_0^{(r)}$ is tested only once 
and the true null corresponds to the unique true winner $\hat{\tau}_{(1)}$.  
Formally:

\begin{theorem}[FWER control of the naive max–statistic test]\label{thm:naive-fwer}
Under Assumptions~\ref{assump1}–\ref{a3}, the selection rule defined above satisfies
\[
\lim_{n\to\infty} 
\P\!\left( \hat{\tau}_{(1)} \in \hat{\mathcal{S}}_{naive} \right)
\ \ge\ 1 - \alpha.
\]
\end{theorem}
The proof of Theorem~\ref{thm:naive-fwer} is provided in Appendix~\ref{app-pf-naive}.

\subsection{Proposed Method} \label{sec-proposed}
\begin{figure}[t]
    \centering
    \vspace{-6pt}
    \begin{minipage}[t]{0.45\textwidth}
        \centering
        \includegraphics[width=\linewidth]{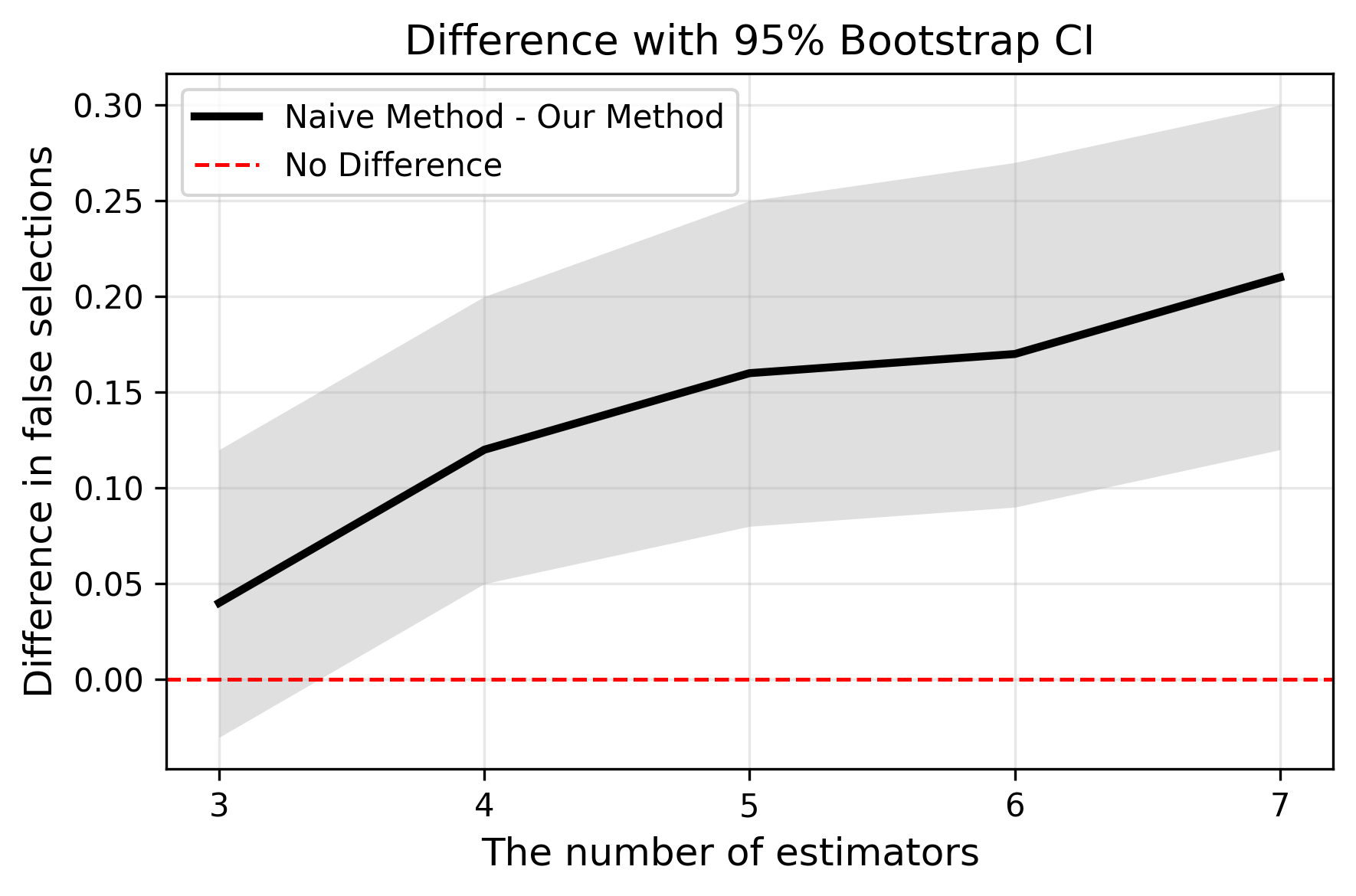}
        \caption{Difference in false selections between the Naive Method and our proposed method (mean $\pm$ 95\% bootstrap confidence interval) under a linear toy model with three competitive and four clearly inferior estimators. See Appendix~\ref{app-toy} for details.}
        \label{fig:naive_ours}
    \end{minipage}
    \hfill
    \begin{minipage}[t]{0.48\textwidth}
        \centering
        \includegraphics[width=\linewidth]{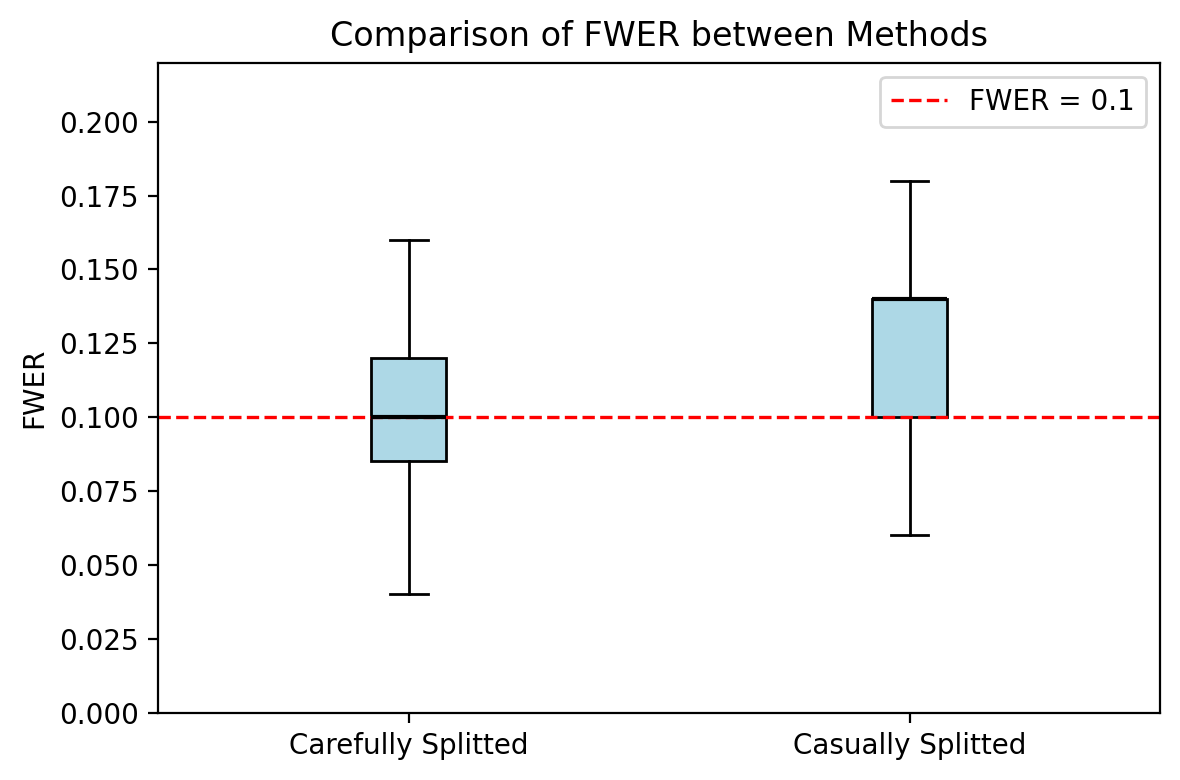}
        \caption{Comparison of FWER under different data-splitting schemes (mean and variability over 100 repetitions) under the linear toy model. The "Casually Splitted" (without two-way splitting) strategy \textbf{inflates FWER}, while our "Carefully Split" scheme ensures proper error control. }
        \label{fig:split_difference}
    \end{minipage}
    \vspace{-6pt}
\end{figure}

\noindent \textbf{Motivation.} While the naive method provides valid error control, it becomes particularly conservative when the candidate set includes irrelevant estimators: the max-statistic treats all candidates symmetrically, causing the critical value to inflate with $p$ regardless of their actual performance. Figure~\ref{fig:naive_ours} illustrates this with a toy experiment—as $p$ increases, the gap in false selections between the naive method and ours widens substantially. This motivates a statistic that reweights estimators by empirical plausibility rather than penalizing them uniformly; further theoretical and empirical comparisons with the naive method are provided in Appendix~\ref{app-power}.

\subsubsection{Testing Rule}
Motivated by Zhang et al.~\citep{zhang2024winners}, we adapt their core 
idea to the setting of multi-estimator causal comparison and construct a test statistic that aggregates pairwise relative errors in a 
way that assigns larger weights to larger error gaps, thereby increasing the power to detect when the estimator to be tested is suboptimal. 
Formally, given an estimator $\hat \tau_r$ to be tested, we calculate the weighted average of relative errors between \(\hat \tau_r\) and other estimators as:
\[
\sum_{i=1}^n Q_{i,r},~~
Q_{i,r} \triangleq  \sum_{s\neq r} \hat\omega_{r,s}\,
        \hat t(Z_i;\hat\tau_r,\hat\tau_s),
\]
where $\hat\omega_{r,s}$ denote some data-driven weight.

To compute the weight $\hat\omega_{r,s}$, we cannot simply apply standard cross-validation as proposed in Zhang et al.~\citep{zhang2024winners}. This is because the relative error estimates $\hat t(Z_i)$ are not fully independent; they are coupled through the shared nuisance function estimation described in Section 2.2, which violates the independence assumption required for valid inference. Neglecting this dependence leads to the failure of FWER control, as empirically demonstrated in the counterexample in Figure~\ref{fig:split_difference}. To address this challenge, we introduce a tailored design---a two-way sample-splitting scheme.
Explicitly, we first partition the dataset into two major folds $A$ and $B$.

\begin{itemize}
    \item Across major folds: on fold $A$, we estimate 
the nuisance components $\hat\eta_A$ and evaluate the relative-error 
quantities $\hat\delta$ on fold $B$; the roles of the folds are then 
swapped to obtain $\hat\delta$ on fold $A$ using $\hat\eta_B$.  
    \item Within each major fold: we perform 
an inner $K$-fold split to compute the exponential weights used in the aggregation 
step. 
\[
\hat{\omega}_{r, s}^{(-i)} \;\propto\; \exp\!\left( \lambda \cdot \hat{\delta}(\hat \tau_r, \hat \tau_s)^{(\mathcal{H} - \{Z_i\})} \right),~~\sum_{s \neq r} \hat \omega_{r, s}^{(-i)} = 1
\]
where \(\lambda\) is a hyper-parameter and \(Z_i \in \text{major fold}\; \mathcal{H}\). 

\end{itemize}
For clarity, we provide a figure illustrating the full data-partitioning scheme in Appendix~\ref{app-partition}.

With the above weights, for a given estimator $\hat{\tau}_r$, let
\[
S_r \triangleq  \sum_{i=1}^n \sum_{s \neq r} \hat \omega_{r,s}^{(-i)} 
\hat{t}(Z_i; \hat{\tau}_r,\hat{\tau}_s;\hat\eta_{\text{opp}}) = \sum_{i=1}^n  Q_{i, r}
\]
Intuitively, $S_r$ 
summarizes how large the estimation error of $\hat{\tau}_r$ is compared with other \(\hat \tau_s\) on the test dataset, with larger value indicating stronger evidence that \(\hat \tau_r\) is not the best performer.
Therefore, we reject the null hypothesis that \(\hat \tau_r\) is the winner within the group if we observe 
\[
\frac{1}{\sqrt{n} \hat \sigma_r} {S}_r > c_\alpha
\] 
where \(\hat \sigma_r^2 \triangleq \Var(Q_{1, r})\)~\cite{zhang2024winners}. We will show later that the critical value \(c_\alpha\) is in fact the standard normal distribution \(1-\alpha\)-quantile \(z_{1-\alpha}\).
The full procedure, including the construction of cross-fitted summaries, 
exponential weighting, and the standardized test statistic, is detailed in 
Algorithm~\ref{alg1}. 
\vspace{-4pt}
\begin{algorithm}[t!]
\caption{Proposed Method: Cross-Fitted Pairwise Exponentially Weighted Confidence Set}
\label{alg1}
\KwIn{Data $Z$, outer split into two major folds $A,B$, inner fold number $K$, significance level $\alpha$, weighting parameter $\lambda$}
\KwOut{Confidence set $\hat{\mathcal{S}}$}
\BlankLine
Initialize $\hat{\mathcal{S}}\leftarrow \varnothing$\;
Train nuisances $\hat\eta_A$ on $A$, $\hat\eta_B$ on $B$\;

\For{estimator index $r \in [p]$}{
  $S_r \leftarrow 0$\;
  
  \For{major fold $\mathcal{H} \in \{A,B\}$}{
    $\eta_{\text{opp}} \gets \hat\eta_A$ if $\mathcal{H}=B$, else $\hat\eta_B$\;
    
    \For{$Z_i \in \mathcal{H}$}{
    
      \For{$s \neq r$}{
        Compute $\hat{t}(Z_i; \hat{\tau}_r, \hat{\tau}_s; \eta_{\text{opp}}) $\;
      }
      $\hat{t}(Z_i; \hat{\tau}_r) 
= 
\underbrace{\left(
\hat{t}(Z_i; \hat{\tau}_r, \hat{\tau}_1), \dots, 
\hat{t}(Z_i; \hat{\tau}_r, \hat{\tau}_p)
\right)^\top}_{p-1 \text{ components}} $\;
    }
    Split $\mathcal{H}$ into $K$ inner folds $\{I_v^{(\mathcal{H})}\}_{v=1}^K$\;
    
    \For{inner fold $v \in [K]$}{
      Compute out-of-fold mean summaries $\hat\delta^{(-v,\mathcal{H})} = \frac{2K}{(K-1)|Z|} \sum_{Z_i \in \mathcal{H} / I_v^{(\mathcal{H})}} \hat{t}(Z_i; \hat{\tau}_r) $\\
      Define weights $\hat\omega^{(-v,\mathcal{H})}_{r,s} \propto \exp(\lambda \hat\delta^{(-v,\mathcal{H})}_s)$, 
      with $\sum_{s \neq r}\hat\omega^{(-v,\mathcal{H})}_{r,s}=1$\;
      
      \For{sample $Z_i \in I_v^{(\mathcal{H})}$}{
        Calculate weighted competitor
        \(
        Q^{(\mathcal{H})}_{i,r} = \sum_{s \neq r} \hat\omega^{(-v,\mathcal{H})}_{r,s}\,
\hat\delta(Z_i;\hat\tau_r,\hat\tau_s;\eta_{\text{opp}})
        \)\\
        Update accumulator $S_r \leftarrow S_r + Q^{(\mathcal{H})}_{i,r}$\;
      }
    }
  }
  Estimate the standard deviation $\hat\sigma_r>0$\;
  
  \If{$\frac{1}{\sqrt{n} \hat \sigma_r} S_r < z_{1-\alpha}$}{
    add $r$ to $\hat{\mathcal{S}}$\;
  }
}
\vspace{-4pt}
\end{algorithm}

\textbf{Choice of the hyperparameter $\lambda$.} The value of $\lambda$
governs the trade-off between power and validity: larger values increase power but must satisfy the constraints imposed later (Theorems~\ref{thm2} and \ref{thm3}). Inspired by \cite{zhang2024winners}, we adopt an adaptive selection strategy that begins with a small initial $\lambda$ and increases it iteratively, stopping as soon as the empirical first-order stability condition (Theorem~\ref{thm2}) is violated.

\textbf{Careful splitting matters.} 
The two-way sample-splitting scheme plays a crucial role in maintaining valid inference. As empirically demonstrated in Figure~\ref{fig:split_difference}, failing to strictly decouple nuisance estimation from weight learning results in a significant inflation of the FWER. This contrast highlights the necessity of our proposed two-way sample splitting for reliable error control.

\subsubsection{FWER Control} 
{We now state our main theoretical guarantee. Under mild conditions on the nuisance estimators,
the proposed selection procedure controls the familywise error rate.}
\begin{theorem}[FWER control of the cross-fitted exponentially weighted test]\label{thm:ours-fwer}
Under Assumptions~\ref{assump1}–\ref{a4}, the selection rule induced by
Algorithm~\ref{alg1} (denoted as \(\hat{\mathcal{S}}_{ours}\)) satisfies
\[
\lim_{n\to\infty} 
\P\!\left( \hat{\tau}_{(1)} \in \hat{\mathcal{S}}_{ours}\right)
\ \ge\ 1 - \alpha.
\]
\end{theorem}
\textbf{Proof.} The quantities $Q_{i,r}$ are not fully independent across $i$. The proof proceeds via a stability-based CLT (Proposition~\ref{thm:stableCLT}), after verifying that our test statistic satisfies the required first- and second-order stability conditions (Theorems~\ref{thm2}–\ref{thm3}).
We first introduce the necessary stability operator.
\begin{definition}
    For distinct $j,l \in [n]$, dataset \(\mathbf{Z} = \{Z_i\}_{i=1}^n\) and a map \(K\), we define the (stability) operator of \(K\) as follows:
\[
\nabla_{j} K := {K}(\mathbf{Z}) - {K}( \mathbf{Z}^{j}),
\]
\[
\nabla_{l} \nabla_{j} K := {K}(\mathbf{Z}) - {K}(\mathbf{Z}^{j}) 
- \left\{{K}(\mathbf{Z}^{l}) - {K}(\mathbf{Z}^{j,l}) \right\},
\]
where the perturbed data sets are defined as $\mathbf{Z}^{j}$ replacing the sample $Z_{j}$ in $\mathbf{Z}$ by an IID copy.
\end{definition}
Now, the dependent data CLT~\cite{zhang2024winners,austern2020asymptotics} can be summarized as:
\begin{proposition}[Stability-based CLT for globally dependent data]
\label{thm:stableCLT}
Let $\mathbf{Z} = \{Z_i\}_{i=1}^n$ be IID observations and consider a 
collection of statistics $K_i \in [-M,M]$ 
satisfying $\mathbb{E}[K_i \mid \mathbf{Z}^{(-i)}]=0$ and 
$v_n^2 := \mathrm{Var}(K_1)$ with $\liminf_n v_n > 0$.  

Suppose the first- and second-order stability measures satisfy

\begin{minipage}{0.45\textwidth}
    \begin{equation}
    \Delta_1^2 = \max_{i \neq j}\; \mathbb{E}[(\nabla_j K_i)^2] = o(n^{-1}) \label{cond1}
    \end{equation}
\end{minipage}%
\hfill
\begin{minipage}{0.45\textwidth}
    \begin{equation}
    \Delta_2^2 = \max_{i \neq j \neq l}\; \mathbb{E}[(\nabla_l \nabla_j K_i)^2] = o(n^{-2}) \label{cond2}
    \end{equation}
\end{minipage}

Then the standardized sum satisfies
\[
v_n^{-1} n^{-1/2}
\sum_{i=1}^n K_i
\;\xrightarrow{d}\;
\mathcal{N}(0,1).
\]
\end{proposition}

The proposition is actually a re-statment of Theorem 3.10 in Zhang et al.~\citep{zhang2024winners} and Theorem 1 in Austern \& Zhou~\citep{austern2020asymptotics}, so we omit the proof here.

{
To apply the stability-based CLT, we require that the nuisance estimators
satisfy a mild local stability condition (Assumption~\ref{a4}):
replacing a single observation in the training sample should induce only
an $O(n^{-1})$ perturbation in the estimated regression functions, with
mixed second-order perturbations being of smaller order.  
This property holds for a broad class of smooth ERM-based learners,
including regularized linear or generalized linear models and kernel ridge
regression, and is formally justified in Appendix~\ref{appendix:local-stability}.
Moreover, in the experiment section we demonstrate that the test
statistics remain the desired properties even when the nuisance
components are fitted using black-box methods such as neural networks, as long as appropriate stability-enhancing techniques are carefully applied.
}

Turning back to our problem, it suffices to show that our choice of \(K_i\) does satisfy the first and second order stability. Under Assumption~\ref{a1}-~\ref{a4}, the following theorems hold.
\begin{theorem}[First Order Stability]\label{thm2}
Let $\hat \tau_r$ be the estimator of interest and $j \in \mathbf{Z}$ and $i \notin \text{inner fold} \; I_{v_j}$ be two sample indices. Define
\(
K_{j,r}(\mathbf{Z}) = Q_{j,r} - \E \!\left[Q_{j,r} \mid Z^{(-j)} \right].
\)
$M>0$ is the upper bound of  $\hat{t}(Z_1; \hat{\tau}_r)-\E[\hat{t}(Z_1; \hat{\tau}_r)]$, then for some universal constant $C>0$,
\begin{equation}\label{eq-first-stability}
\max_{i,j}\|\nabla_i K_{j,r}\|_2  \le C \lambda M^2 n^{-1}
\end{equation}
for sufficiently large \(n\).
In particular, when $\lambda = o(\sqrt{n})$, we have
\(
\max_{i,j}\|\nabla_i K_{j,r}\|_2 = o(n^{-1/2}).
\)
\end{theorem}

\vspace{-6pt}
\begin{figure*}[t]
\centering
\begin{minipage}[t]{0.32\textwidth}
    \centering
    \includegraphics[width=\textwidth]{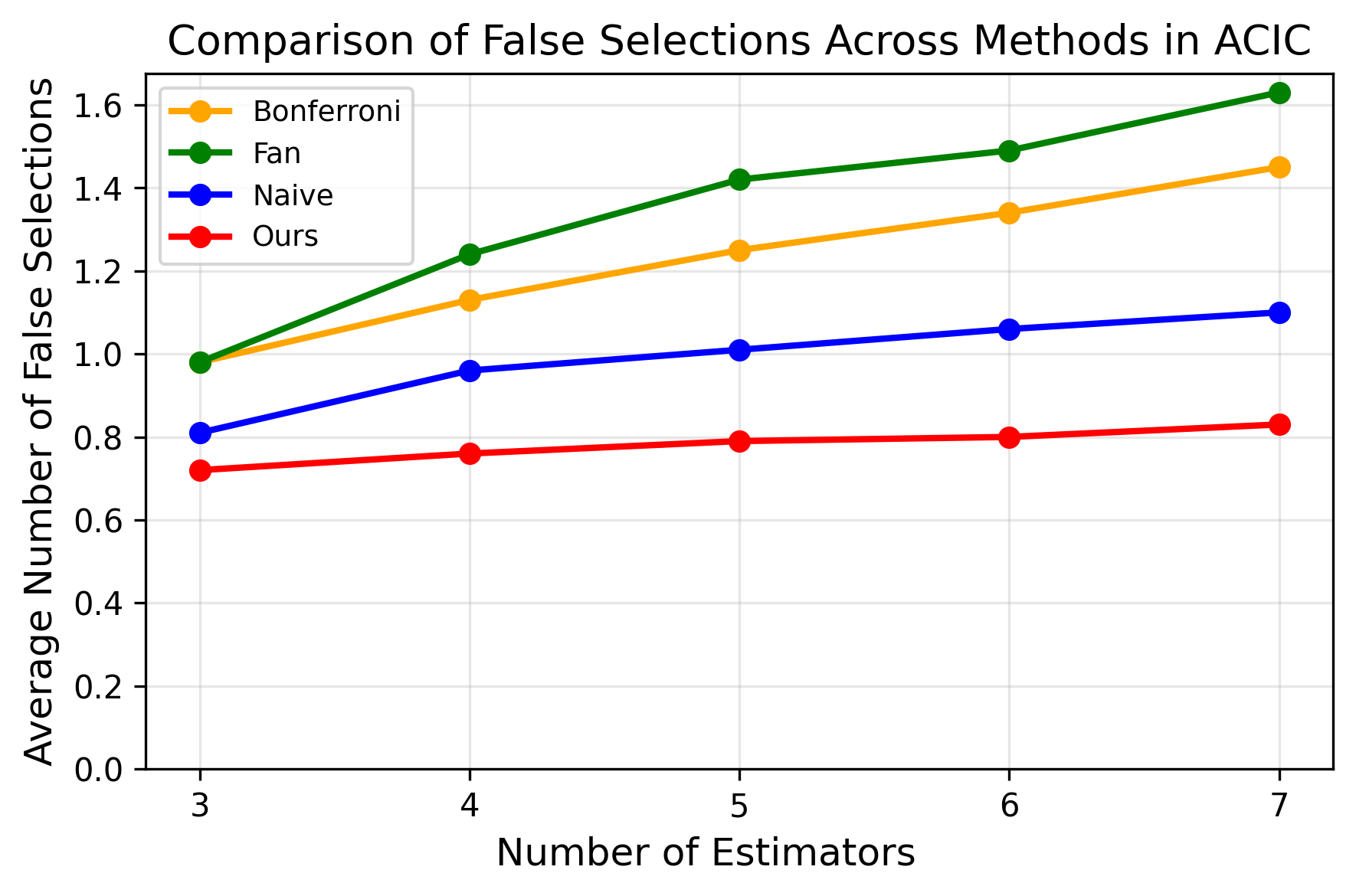}
    \caption*{\textbf{(a)} ACIC dataset}
\end{minipage}
\hfill
\begin{minipage}[t]{0.32\textwidth}
    \centering
    \includegraphics[width=\textwidth]{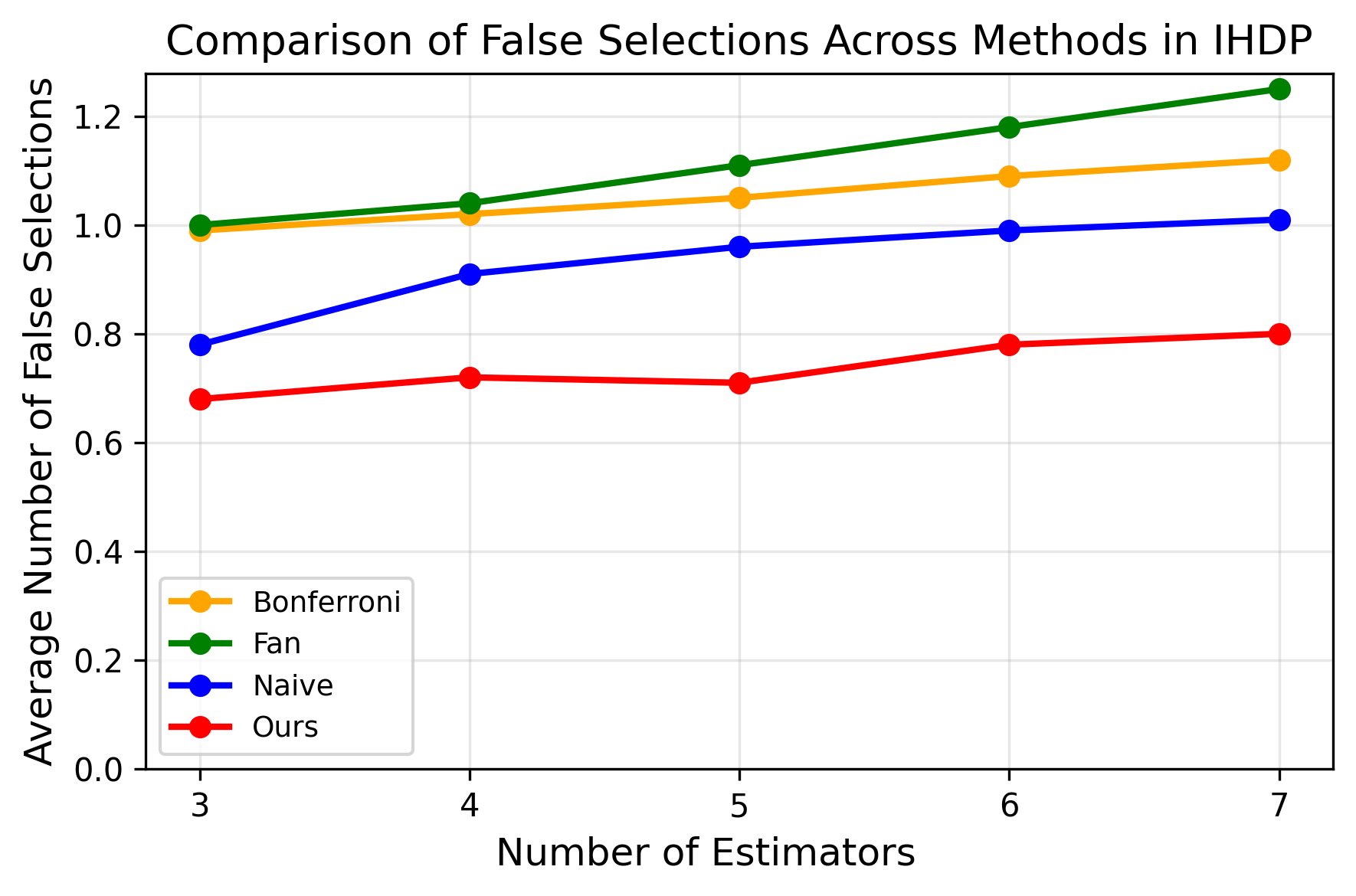}
    \caption*{\textbf{(b)} IHDP dataset}
\end{minipage}
\hfill
\begin{minipage}[t]{0.32\textwidth}
    \centering
    \includegraphics[width=\textwidth]{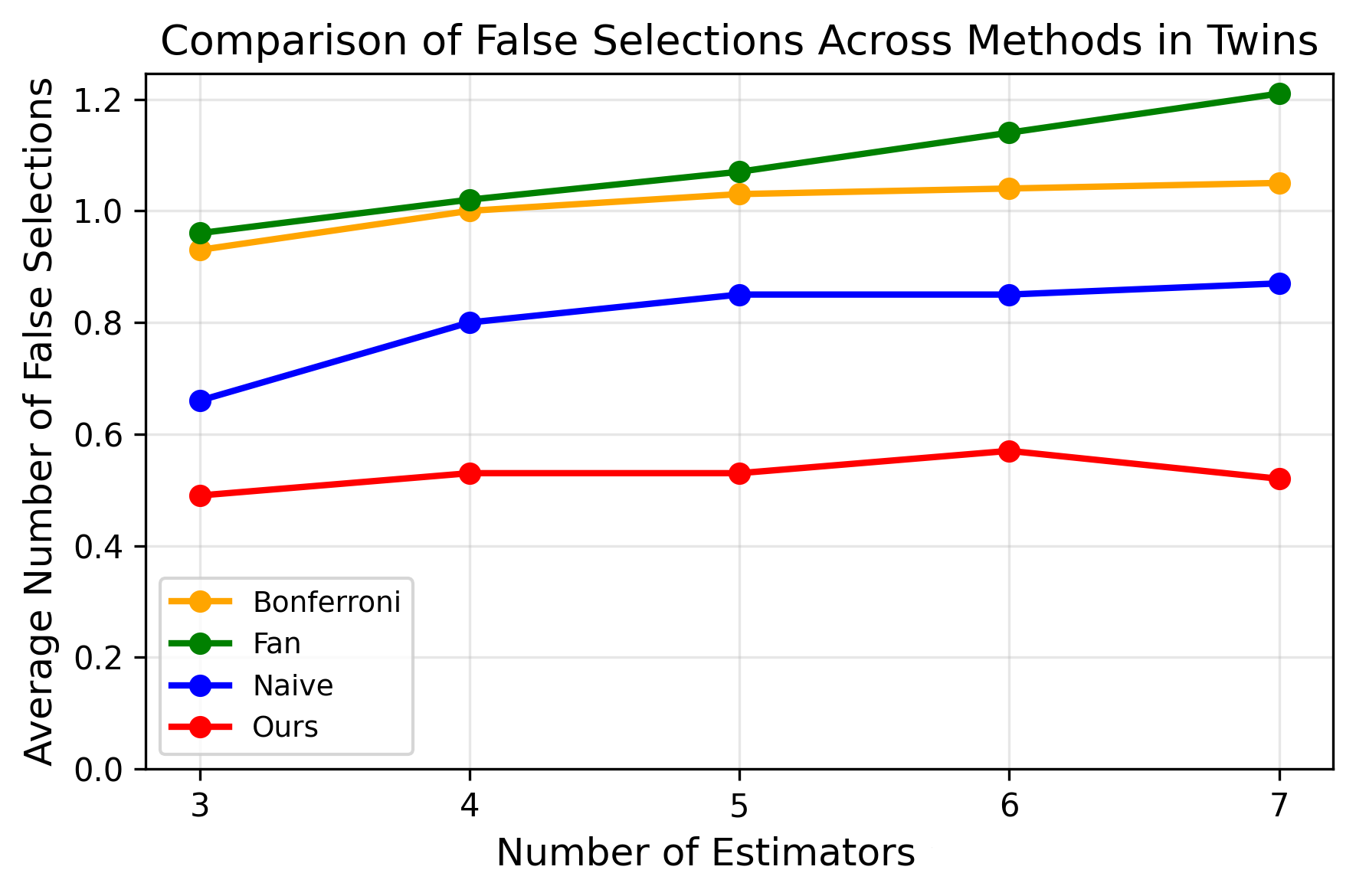}
    \caption*{\textbf{(c)} Twins dataset}
\end{minipage}

\vspace{4pt}
\caption{
Average number of false selections across different numbers of estimators~$p$ 
on three benchmark datasets. 
Our method consistently yields fewer false selections than Naive, 
ranking inference, and Bonferroni baselines.
}
\label{fig:K_datasets}
\vspace{-6pt}
\end{figure*}

\begin{theorem}[Second Order Stability]
\label{thm3} Let $\hat \tau_r$ be the estimator of interest and let $j \in \mathbf{Z}$ and $i \notin \text{inner fold} \; I_{v_j}$ be two sample indices. Under the same conditions of Theorem~\ref{thm2}, the following holds,
\begin{equation}
    \max _{i, j, k}\left\|\nabla_i \nabla_k K_{j,r}\right\|_2 \leq C\lambda^2 M^3 n^{-2}
    \end{equation}
for large enough $n$, a universal constant $C$. In particular, when $\lambda = o(\sqrt{n})$, we have $\max_{i,j,k,r}\norm{\nabla_i \nabla_k K_{j,r}}_2 = o(n^{-1})$.
\end{theorem}
The proofs are provided in Appendix~\ref{app-pf1} and Appendix~\ref{app-pf2}, respectively. {Combining the results above completes the proof of Theorem~\ref{thm:ours-fwer}.}

\vspace{-4pt}
\section{Experiments}\label{sec:experiments}

\subsection{Experimental Setup}
\vspace{-4pt}

\noindent \textbf{Datasets and Preprocessing.} 
Following previous studies~\cite{pmlr-v70-shalit17a, louizos2017causal, chauhan2024dynamic}, we evaluate our method on three benchmark datasets widely used in causal inference: \textbf{ACIC 2016}, \textbf{IHDP}, and \textbf{Twins}. ACIC 2016~\cite{dorie2019automated} is a semi-synthetic dataset with 4802 samples and 55 covariates, derived from the Collaborative Perinatal Project~\cite{niswander1972women}. IHDP contains 747 samples (139 treated, 608 control) with 25 covariates, targeting the effect of specialist home visits on infant cognitive outcomes. Twins~\cite{10.1093/qje/120.3.1031} comprises 5271 same-sex twin birth records with 28 covariates. More dataset details are provided in Appendix~\ref{app-dataset}.

\noindent \textbf{Estimators Compared.} To create multiple candidate HTE estimators, we construct a collection of seven Causal Forests~\cite{wager2018estimation} with different values of the number of trees (\texttt{n\_estimators}) and the maximum tree depth (\texttt{max\_depth}), detailed in Appendix~\ref{app-candidate-details}. These estimators are intentionally made similar so that the comparison task is difficult. To further assess robustness across diverse estimator types, we conduct a supplementary experiment with heterogeneous candidates in Appendix~\ref{app-diverse}.


\noindent \textbf{Relative error construction.} {We adopt the DragonNet architecture~\cite{shi2019adapting} for the nuisance estimation in the relative error construction. We also incorporate several stability-oriented refinements, including exponential moving average (EMA) of parameters~\cite{tarvainen2018meanteachersbetterrole}, a warm-up training phase~\cite{goyal2018accuratelargeminibatchsgd}, additive Gaussian noise regularization~\cite{vincent2008extracting}, and layer normalization~\cite{ba2016layer}.} Although it remains a black-box model with no theoretical guarantee of prediction stability, our simulation studies (see Appendix~\ref{app-nuisance}) indicate that using this set of nuisance estimators still satisfy the conditions required by our procedure.

\noindent \textbf{Baselines for Estimator Selection.}  
We compare our method against three alternatives:
    (i) the naive method proposed in Section~\ref{sec-naive}; 
    (ii) Bonferroni correction method~\cite{weisstein2004bonferroni};
    (iii) the ranking inference method provided by Fan et al.~\citep{Fan02012025}.
All methods target the same hypothesis of identifying the best-performing estimator among the $p$ candidates.

\noindent \textbf{Evaluation Metrics.} We report the familywise error rate (FWER) and the average number of incorrect selections per experiment as our evaluation metrics. The significance level is set to 0.10, and each dataset is evaluated over 100 independent repetitions.
\begin{table*}[t]
\centering
\caption{Familywise error rate (FWER) and average number of false selections (ANFS) 
on three benchmark datasets (mean $\pm$ standard error over 100 repetitions). 
The best results are bolded.}
\vspace{-8pt}
\label{tab:fwer}
\resizebox{0.8\textwidth}{!}{
\begin{tabular}{l|cc|cc|cc}
\toprule
\multicolumn{1}{l|}{} &
\multicolumn{2}{c|}{\textbf{\textit{ACIC}}} &
\multicolumn{2}{c|}{\textbf{\textit{IHDP}}} &
\multicolumn{2}{c}{\textbf{\textit{Twins}}} \\
\cmidrule(lr){2-3} \cmidrule(lr){4-5} \cmidrule(lr){6-7}
\textbf{Method} 
& \textbf{FWER} & \textbf{ANFS}
& \textbf{FWER} & \textbf{ANFS}
& \textbf{FWER} & \textbf{ANFS} \\
\midrule
Naive     & 0.03  & 1.10 $\pm$ 0.04 & 0.02  & 1.01 $\pm$ 0.03 & 0.01 & 0.88 $\pm$ 0.04 \\
Bonferroni     & 0.01  & 1.45 $\pm$ 0.05 & 0.01  & 1.12 $\pm$ 0.03 & 0 & 1.05 $\pm$ 0.02 \\
Ranking Inference    & 0 & 1.63 $\pm$ 0.06 & 0.01  & 1.23 $\pm$ 0.04 & 0 & 1.18 $\pm$ 0.04 \\
\midrule
\textbf{Ours} 
& 0.02 & \textbf{0.83 $\pm$ 0.06}
& 0.04 & \textbf{0.80 $\pm$ 0.06}
& 0.02 & \textbf{0.50 $\pm$ 0.05} \\
\bottomrule
\end{tabular}
}
\vspace{-6pt}
\end{table*}
\subsection{Main Results}

\textbf{FWER and Incorrect Selections.}
Table~\ref{tab:fwer} reports the familywise error rate (FWER) and the average number of false selections (ANFS) over 100 replications. While all methods manage to control the FWER, it can be seen clearly that our method substantially reduces ANFS compared with existing baselines across all three datasets.
Notably, the improvement is most pronounced on ACIC2016 and Twins, which may be attributed to the relatively larger sample sizes in these datasets. 

\noindent \textbf{Effect of the Number of Estimators $p$.}  As shown in the toy model provided in Section~\ref{sec-proposed}, the number of the estimators to be tested may significantly influence the performance of the evaluation methods. To demonstrate this, we vary \(p\) within our estimator list while maintaining the true winner.
Figure~\ref{fig:K_datasets} shows how ANFS changes as the number of candidate estimators increases from 3 to 7.  
Across all datasets, our method is remarkably insensitive to the number of estimators being compared: as $p$ increases, the number of incorrect selections remains almost unchanged. In contrast, all competing methods deteriorate noticeably as $p$ grows—they not only start off worse than ours, but the performance gap continues to widen. This phenomenon directly reflects the motivation behind the design of our procedure.

\begin{figure}[t!]
    \centering

    \begin{minipage}[t]{0.45\textwidth}
        \centering
        \includegraphics[width=\textwidth]{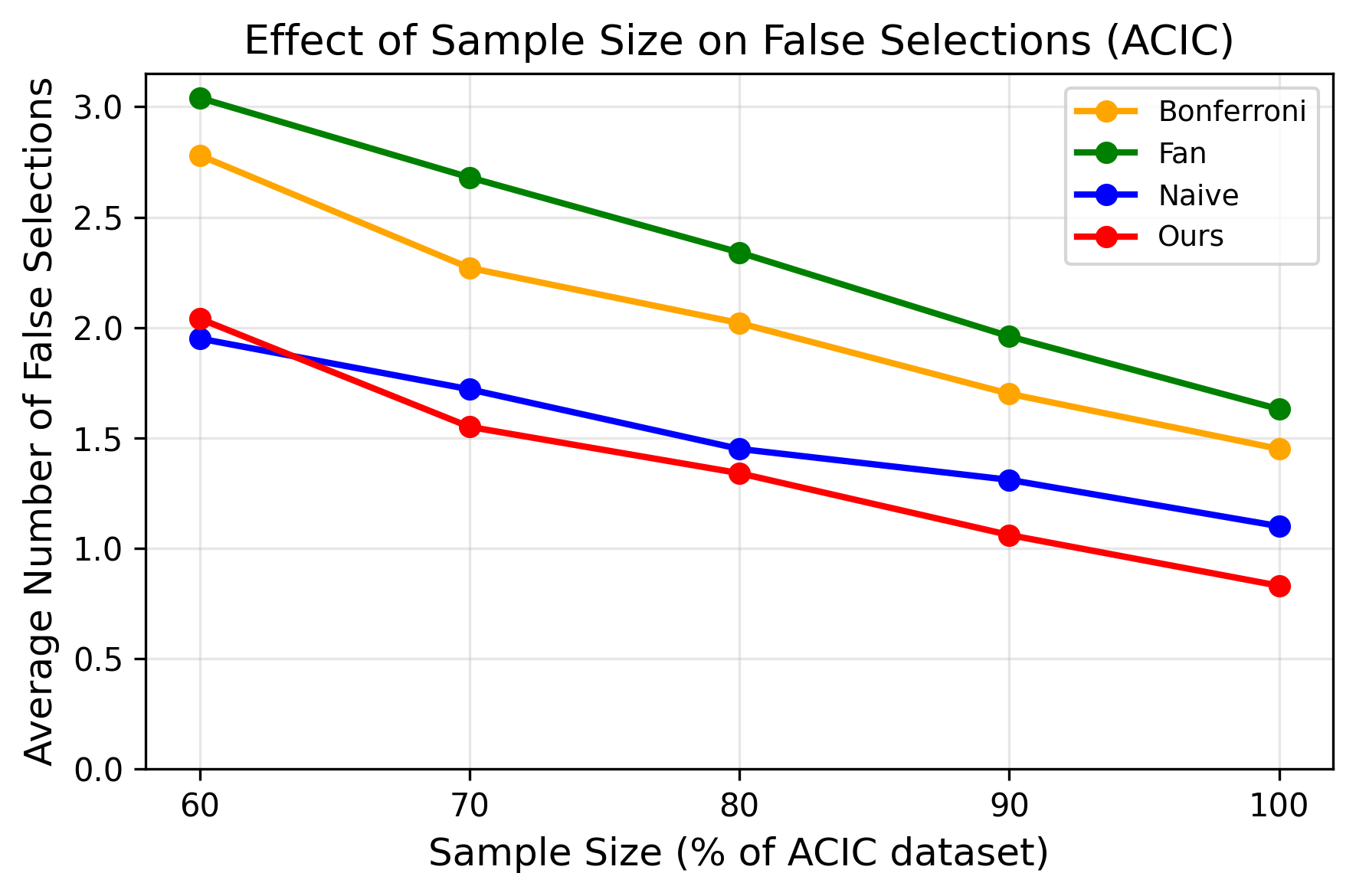}
        \caption*{\textbf{(a)} ACIC2016 dataset}
    \end{minipage}
    \hfill
    \begin{minipage}[t]{0.45\textwidth}
        \centering
        \includegraphics[width=\textwidth]{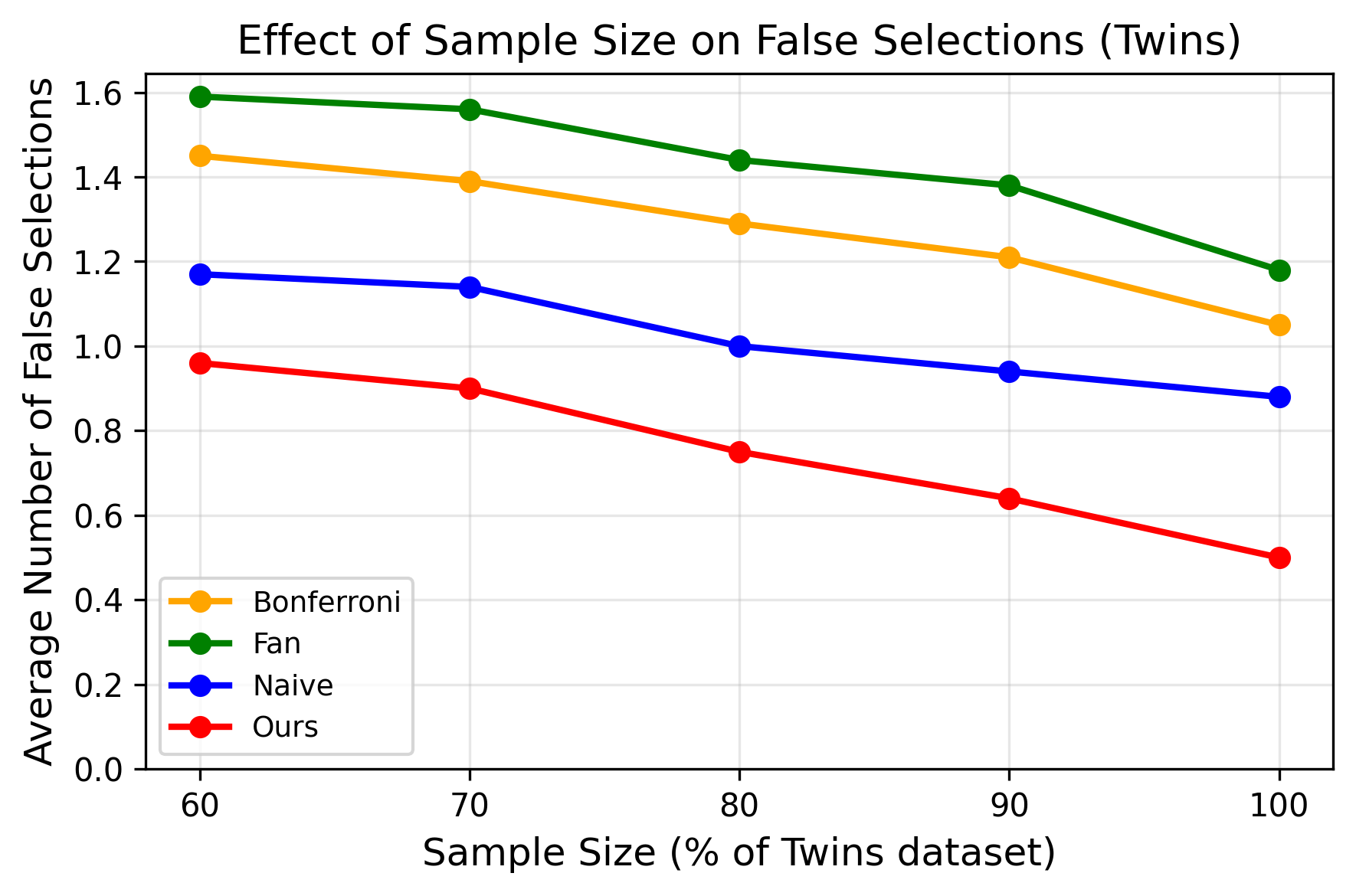}
        \caption*{\textbf{(b)} Twins dataset}
    \end{minipage}
    \caption{
        Effect of sample size on the average number of false selections.
        Across both ACIC2016 and Twins, our method benefits more from additional samples
        and consistently achieves the lowest number of false selections across all regimes.
    }
    \label{fig:sample_size}
    \vspace{-6pt}
\end{figure}

\noindent \textbf{Effect of Sample Size.} A natural question is whether 
our method improves significantly as the test set grows. We vary the sample 
size from 60\% to 100\% of the original datasets; results on ACIC2016 and 
Twins are shown in Figures~\ref{fig:sample_size} (a) and (b). All methods 
improve with more data, but our method shows the steepest improvement curve 
and maintains the largest advantage in low-sample regimes.

\FloatBarrier
\vspace{-4pt}
\section{Conclusion}
\vspace{-4pt}

{In this paper, we formulate the problem of selecting the best estimator among multiple HTE estimators as an inferential question of the argmin of unknown risks, and propose a ground-truth-free procedure for selecting the best one. Our cross-fitted exponentially weighted test statistic leverages stability-based CLTs to guarantee asymptotic FWER control under mild nuisance-estimator conditions. Empirically, the method achieves reliable FWER control while substantially reducing false selections across benchmark datasets, with advantages most pronounced when the candidate pool is large or estimators are highly similar.
}
\bibliographystyle{unsrt}
\bibliography{example_paper}

\clearpage
\appendix
\appendix
In \Cref{app-notation}, we provide a table of notation. In \Cref{appe:sec:related.work}, we discuss related work. In \Cref{appe:sec:algorithm}, we present the algorithm for the naive method. In \Cref{app-pf-naive,app-pf1,app-pf2}, we collect the proofs of the theoretical results in the main text. In \Cref{appendix:local-stability}, we justify the stability assumption. In \Cref{appe:sec:experiment}, we provide additional experimental details.

\section{Notation Summary}\label{app-notation}
\begin{table}[ht]
\centering
\caption{Notation summary.}
\label{tab:notation}
\begin{tabular}{ll}
\toprule
\textbf{Symbol} & \textbf{Meaning} \\
\midrule
$W$ & Binary treatment variable \\
$X$ & Pre-treatment covariates \\
$Y$ & Outcome \\
$Z$ & A combination of $(X, Y, W)$\\
$\tau(x)$ & Individual treatment effect \\
$e(x)$ & Propensity score \\
$\mu_a(x)$ & Outcome regression function, i.e., $\mu_a(x) = \mathbb{E}[Y \mid X=x, W=a]$ for $a=0,1$ \\
$\delta(\hat{\tau}_1, \hat{\tau}_2)$ & (real) Relative error between estimator $\hat{\tau}_1$ and $\hat{\tau}_2$ \\
$\hat{t}(Z_i, \hat{\tau}_r, \hat{\tau}_t; \hat{\eta})$ & Estimated relative error between $\hat{\tau}_r$ and $\hat{\tau}_t$ on $Z_i$ (given nuisance estimator $\hat \eta$)\\
$\eta_A/\eta_B$ & nuisance estimators trained on Fold A/B\\
\bottomrule
\end{tabular}
\end{table}

\section{Related Work}\label{appe:sec:related.work}
\subsection{Evaluation of HTE Estimators} A wide range of evaluation frameworks have been proposed for heterogeneous and conditional treatment effect estimators, reflecting the fact that individual-level causal effects are never directly observable. Early work formalized proxy criteria such as PEHE~\cite{Hill01012011}, now standard in semi-synthetic benchmarks, though subsequent analyses~\cite{crabbe2022benchmarking} noted that PEHE captures only pointwise prediction accuracy and may incentivize unrealistic overfitting. To address settings where individual treatment effects cannot be recovered, Shalit et al.~\citep{shalit2017estimating} introduced policy risk, which evaluates how well a model’s treatment recommendations perform relative to an oracle and has been used extensively in applications such as the Jobs dataset. Gao~\citep{Gao2024} introduced a relative-error measure for pairwise comparison of two
estimators, yielding robust conclusions. Guo et al.~\citep{guo2025relativeerrorbasedevaluationframework} further extend
this robustness to a broader setting. Finally, complementing empirical metrics, theoretical analyses~\cite{nie2021quasi, alaa2018limits} characterize the fundamental limits of HTE estimation and identify conditions under which learners can achieve quasi-oracle performance. 

\subsection{Argmin Inference}
Research on argmin inference dates back to early work on identifying minimal elements in multivariate systems~\cite{gibbons1977action}, with later refinements relying on restrictive assumptions such as known marginal distributions or independence~\cite{Futschik}. Modern approaches have since relaxed these conditions, including general confidence-set constructions based on pairwise comparisons~\cite{mogstad2024inference}, bootstrap-based procedures such as~\cite{hansen2011model}, and recent sample-splitting strategies tailored to argmin selection~\cite{zhang2024winners,kim2025locally}. Conceptually, argmin inference is tightly connected to ranking inference, since identifying the minimizer can be framed as determining whether an index achieves rank one. The ranking-inference literature provides complementary tools for constructing confidence sets for discrete ranks~\cite{Fan02012025, goldstein1996league, xie2009confidence}, though sometimes they do not transfer directly to the argmin setting.

\section{Additional Algorithm}\label{appe:sec:algorithm}
Algorithm~\ref{alg:naive} summarizes the naive method.

\begin{algorithm}[ht]
\caption{Naive Max–Statistic Selector}
\label{alg:naive}
\KwIn{Estimators $\{\hat{\tau}_1,\dots,\hat{\tau}_K\}$, two-fold splits $(A, B)$, significance level $\alpha$, number of resamples $\tilde{B}$}
\KwOut{Estimated winner set $\hat{\mathcal{S}}$}
\BlankLine
Initialize $\hat{\mathcal{S}} \leftarrow \varnothing$\;
\For{fold $M \in \{A, B\}$}{
  Train nuisance estimator $\hat{\eta}_M$\;
}

\For{estimator index $m \in [K]$}{
  \For{$s \in \mathcal{I}_m$}{
    Compute $\hat{\delta}(\hat{\tau}_m, \hat{\tau}_s)
    = \frac{1}{2}\!\left[
    \frac{1}{|A|}\!\sum_{Z_i \in A}\!\hat{t}(Z_i;\hat{\tau}_m,\hat{\tau}_s;\hat{\eta}_B)
    + 
    \frac{1}{|B|}\!\sum_{Z_i \in B}\!\hat{t}(Z_i;\hat{\tau}_m,\hat{\tau}_s;\hat{\eta}_A)
    \right]$\;
  }
  Stack $\hat{\delta}_m = \{\hat{\delta}(\hat{\tau}_m, \hat{\tau}_s)\}_{s\in\mathcal{I}_m}$ and estimate $\hat{\Sigma}_m$ from vector scores\;
  
  \For{resample index $b \in [\tilde B]$}{
    Draw $G^{(b)} \sim \mathcal{N}(0,\hat{\Sigma}_m)$\;
    Compute 
    \(
    M^{(b)} = 
    \max_{s \in \mathcal{I}_m}
    \frac{G^{(b)}_s}{\sqrt{(\hat{\Sigma}_m)_{ss}}}
    \)\;
  }
  Set $c_{1-\alpha}^{(m)} \leftarrow$ empirical $(1-\alpha)$-quantile of $\{M^{(b)}\}_{b=1}^B$\;
  
  Compute 
  \(
  S_{m,s} = 
  \frac{\hat{\delta}(\hat{\tau}_m, \hat{\tau}_s)}{\sqrt{(\hat{\Sigma}_m)_{ss}}},
  \quad s \in \mathcal{I}_m
  \)\;
  
  Compute $S_m^{\max} = \max_{s\in\mathcal{I}_m} S_{m,s}$\;
  
  \If{$S_m^{\max} \le c_{1-\alpha}^{(m)}$}{
    Add $m$ to $\hat{\mathcal{S}}$\;
  }
}
\Return{$\hat{\mathcal{S}}$}\;
\end{algorithm}

\section{Data Partitioning Scheme} 
Figure~\ref{fig:split} provides a illustration of the proposed data partitioning scheme.
\label{app-partition}
\begin{figure}[t!]
\centering
\scalebox{0.7}{
\begin{tikzpicture}[
    thick,
    font=\Large
]

\def\w{12}  
\def\h{1.2} 

\draw[fill=colorB] (0, 0)   rectangle (\w, \h)   node[midway] {$B_K$};
\draw[fill=colorB] (0, \h)  rectangle (\w, 2*\h) node[midway] {$\dots$};
\draw[fill=colorB] (0, 2*\h) rectangle (\w, 3*\h) node[midway] {$B_2$};
\draw[fill=colorB] (0, 3*\h) rectangle (\w, 4*\h) node[midway] {$B_1$};

\draw[fill=colorA] (0, 4*\h) rectangle (\w, 5*\h) node[midway] {$A_K$};
\draw[fill=colorA] (0, 5*\h) rectangle (\w, 6*\h) node[midway] {$\dots$};
\draw[fill=colorA] (0, 6*\h) rectangle (\w, 7*\h) node[midway] {$A_2$};
\draw[fill=colorA] (0, 7*\h) rectangle (\w, 8*\h) node[midway] {$A_1$};

\node at (\w/2, 8*\h + 0.5) {Whole dataset};

\node at (-1.5, 6*\h) {Fold A};
\node at (-1.5, 2*\h) {Fold B};

\end{tikzpicture}
}
\caption{The dataset is first split into Fold A (top) and Fold B (bottom). 
Each half is then partitioned into $K$ subfolds.}
\label{fig:split}
\end{figure}

\section{Power Analysis of the Naive Method} \label{app-power}
\subsection{Theoretically}

\noindent\textbf{Setup.} Consider $p$ candidate estimators where $\hat{\tau}_1$ is the true winner, $\hat{\tau}_2$ is a competitive runner-up, and the remaining $p-2$ estimators are clearly inferior:
\[
\delta(\hat{\tau}_1, \hat{\tau}_2) = -\epsilon < 0, \qquad \delta(\hat{\tau}_1, \hat{\tau}_k) = -\Delta \ll -\epsilon, \quad k \geq 3.
\]

\noindent\textbf{Naive method.} When testing $\hat{\tau}_2$ as the candidate winner, the naive critical value $c_{1-\alpha}^{(2)}$ is the $(1-\alpha)$-quantile of $\max_{s \neq 2} G_s$, where $(G_1, \ldots, G_{s\neq 2}) \sim N(0, \Sigma_2)$ under the null. Since this maximum is taken over all $p-1$ competitors regardless of their competitiveness, $c_{1-\alpha}^{(2)}$ is strictly larger than $z_{1-\alpha}$ and grows with $p$. In the independent case, $c_{1-\alpha}^{(2)} = \Phi^{-1}((1-\alpha)^{1/(p-1)})$, which diverges as $p \to \infty$; the dependent case inherits the same qualitative behavior for any fixed positive-definite $\Sigma_2$.

\noindent\textbf{Proposed method.} The exponential weight assigned to competitor $k \geq 3$ satisfies
\[
\hat{\omega}_{2,k}^{(-i)} \propto \exp\!\left(\lambda \hat{\delta}(\hat{\tau}_2, \hat{\tau}_k)\right) \to 0 \quad \text{in probability as } \Delta \to \infty,
\]
since $\delta(\hat{\tau}_2, \hat{\tau}_k) = \epsilon - \Delta \to -\infty$. The test statistic therefore concentrates on the single relevant comparison against $\hat{\tau}_1$:
\[
\frac{S_2}{\sqrt{n}\,\hat{\sigma}_2} \xrightarrow{d} N\!\left(\frac{\epsilon}{\sigma_{2,1}}, 1\right),
\]
and the critical value remains $z_{1-\alpha}$, the standard single-comparison threshold.

\noindent\textbf{Power comparison.} The power to correctly reject $H_0^{(2)}$ (excluding the runner-up $\hat{\tau}_2$ from $\hat{\mathcal{S}}$) satisfies:
\[
\Phi\!\left(\frac{\epsilon\sqrt{n}}{\sigma_{2,1}} - c_{1-\alpha}^{(2)}\right) \leq \Phi\!\left(\frac{\epsilon\sqrt{n}}{\sigma_{2,1}} - z_{1-\alpha}\right),
\]
with the gap $c_{1-\alpha}^{(2)} - z_{1-\alpha}$ growing without bound as $p\to\infty$. Thus for any fixed $\epsilon > 0$ and sufficiently large $p$, the naive method fails to reject $H_0^{(2)}$, leaving $\hat{\tau}_2$ as a false selection, while our method's power remains bounded away from zero.

\subsection{Empirically}
Table~\ref{tab:power} tracks the naive method's critical value $c_{1-\alpha}$ and selected set as irrelevant estimators are added to the pool, using a simulated dataset. It explicitly demonstrates that introducing irrelevant estimators inflates the naive threshold, leading to severe power loss. The data simulation process can be found in Appendix~\ref{app-toy}.

\begin{table}[h]
\centering
\caption{True errors of candidate estimators.}
\label{tab:true-errors}
\begin{tabular}{cccc}
\toprule
Estimator & True Error & Rank & Note \\
\midrule
$M_1$ & 0.01002 & 1 & True Winner \\
$M_2$ & 0.01097 & 2 & Competitive \\
$M_3$ & 0.01116 & 3 & Competitive \\
$M_4$ & 0.09762 & 4 & Irrelevant \\
$M_5$ & 0.09961 & 5 & Irrelevant \\
$M_6$ & 0.09986 & 6 & Irrelevant \\
$M_7$ & 0.10151 & 7 & Irrelevant \\
\bottomrule
\end{tabular}
\end{table}

\begin{table}[h]
\centering
\caption{Performance of the naive method as irrelevant estimators are added. The critical value $c_{1-\alpha}$ inflates monotonically, causing the selected set to grow.}
\label{tab:power}
\begin{tabular}{cclcc}
\toprule
Pool Size ($p$) & Avg.\ $c_{1-\alpha}$ & Candidate Pool & Selected Set $\hat{\mathcal{S}}$ & False Selections \\
\midrule
3 & 1.570 & $\{M_1, M_2, M_3\}$ & $\{M_1, M_2, M_3\}$ & 2 \\
4 & 1.727 & $\{M_1,\ldots,M_4\}$ & $\{M_1, M_2, M_3, M_4\}$ & 3 \\
5 & 1.833 & $\{M_1,\ldots,M_5\}$ & $\{M_1, M_2, M_3, M_4\}$ & 3 \\
6 & 1.918 & $\{M_1,\ldots,M_6\}$ & $\{M_1, M_2, M_3, M_4, M_6\}$ & 4 \\
7 & 1.975 & $\{M_1,\ldots,M_7\}$ & $\{M_1,\ldots,M_7\}$ & 6 \\
\bottomrule
\end{tabular}
\end{table}

\section{Proof of Theorem~\ref{thm:naive-fwer}} \label{app-pf-naive}
\begin{proof}
Without loss of generality, we assume \(\tau_1 \succ \tau_2 \succ \dots \succ \tau_K\), where \(\tau_1\) is the true winner.
\begin{align*}
    \mathbb{P}\left(\tau_1 \notin {\hat{\mathcal{S}}}\right) &= \P \left( \max_{i \neq 1} S_{1,i}> c_{1-\alpha}^{(1)}; \; S_{1,i} = \hat{\delta}(\hat \tau_1, \hat \tau_i) / \sqrt{(\hat \Sigma_1)_{ii}} \right) \qquad (j \triangleq \arg \max S_{1,i}) \\
    &= \P \left(S_{1,j} - \delta(\hat \tau_1, \hat \tau_j) /(\hat \Sigma_1)_{jj} + \delta(\hat \tau_1, \hat \tau_j) /(\hat \Sigma_1)_{jj} > c_{1-\alpha}^{(1)}  \right)\\
    &\approx \P \left(S_{1,j} - \delta(\hat \tau_1, \hat \tau_j) /(\Sigma_1)_{jj} + \delta(\hat \tau_1, \hat \tau_j) /(\Sigma_1)_{jj} > c_{1-\alpha}^{(1)}  \right)\\
    &\leq \P \left(S_{1,j} - \delta(\hat \tau_1, \hat \tau_j) /(\hat \Sigma_1)_{jj} > c_{1-\alpha}^{(1)}  \right) \qquad (\delta(\hat \tau_1, \hat \tau_j)<0)\\
    &= \P \left((\hat \delta(\hat \tau_1, \hat \tau_j) - \delta(\hat \tau_1, \hat \tau_j) )/(\hat \Sigma_1)_{jj} > c_{1-\alpha}^{(1)}  \right) \\
    &= \P \left(G_j / (\Sigma_1)_{jj} > c_{1-\alpha}^{(1)} \right) \qquad (G \overset{d}{=} \hat{\delta}(\hat \tau_1, \cdot) - {\delta}(\hat \tau_1, \cdot))\\
    &\leq \alpha.  
\end{align*}
The equality holds only when all estimators perform identically, 
that is, $\delta(\hat{\tau}_1, \hat{\tau}_j) = 0$ for all $j \neq 1$.
Under the unique winner assumption, this equality cannot occur, 
so the inequality is strict. 
Nevertheless, when the performance gaps between estimators are sufficiently small, 
the bound may be nearly attained in finite samples.
\end{proof}

\section{Proof of Theorem~\ref{thm2}} \label{app-pf1}
\begin{proof}
By definition,
\begin{align}
\nabla_i K_{j,r}
&= K_{j,r}(\mathbf{Z}) - K_{j,r}(\mathbf{Z}^{(i)}) \notag\\
&= \Big(Q_{j,r}^{(M)} -\E[Q_{j,r}^{(M)} \mid Z^{(-j)}]\Big)
   - \Big(Q_{j,r}^{(M),(i)} - \E[Q_{j,r}^{(M),(i)} \mid Z^{(-j),(i)}]\Big).
\end{align}
For simplicity, denote $\hat{t}_s(Z_j) := \hat{t}(Z_j; \hat\tau_r,\hat\tau_s;\eta_{\text{opp}})$. The variance of \(\hat t_s(Z_j)\) is bounded~\cite{Gao2024}, the upper bound denoted as \(v^2\).

\noindent \textbf{Case 1: $i$ and $j$ in the same outer fold but different inner folds.} This case can be proven using a similar strategy as in Zhang et al.~\citep{zhang2024winners}.
In this case $\hat{t}_s(Z_j) = \hat{t}_s^{(i)}(Z_j)$, so the difference arises only from the weights:
\begin{align}
\E &\Big( (Q_{j,r}^{(M)} -\E[Q_{j,r}^{(M)} \mid Z^{(-j)}]) - (Q_{j,r}^{(M),(i)} - \E[Q_{j,r}^{(M),(i)} \mid Z^{(-j),(i)}])\Big)^2 \notag\\
&= \E \Bigg(\sum_{s \neq r} (\hat{\omega}_{r,s} - \hat{\omega}_{r,s}^{(i)}) \big(\hat{t}_s(Z_j) - \E[\hat{t}_s(Z_j) \mid Z^{(B)}]\big)\Bigg)^2 \notag\\
&= \E \Bigg(\sum_{s \neq r} \hat{\omega}_{r,s} \Big(1 - \frac{\hat{\omega}_{r,s}^{(i)}}{\hat{\omega}_{r,s}} \Big)\big(\hat{t}_s(Z_j) - \E[\hat{t}_s(Z_j) \mid Z^{(B)}]\big)\Bigg)^2.
\end{align}

To control the ratio $\hat\omega_{r,s}^{(i)}/\hat\omega_{r,s}$, define $\tilde n = n(1-1/K)/2$ being the sample size of the remained inner folds (which are used to compute \(\hat \omega\)). For any $s\ne r$,
\begin{align}
\frac{\hat\omega_{r,s}^{(i)}}{\hat\omega_{r,s}}
&= \frac{\exp(\lambda \hat\delta_s^{(i)})}{\sum_{t\ne r}\exp(\lambda \hat\delta_t^{(i)})} \cdot \frac{\sum_{t\ne r}\exp(\lambda \hat\delta_t)}{\exp(\lambda \hat\delta_s)} \notag\\
&= \exp(\lambda \tilde n^{-1}(\hat{t}_s(Z_i')-\hat{t}_s(Z_i))) \cdot \frac{\sum_{t\ne r}\exp(\lambda \hat\delta_t)\exp(\lambda \tilde n^{-1}(\hat{t}_t(Z_i)-\hat{t}_t(Z_i')))}{\sum_{t\ne r}\exp(\lambda \hat\delta_t)} \notag\\
&\le \exp\!\Big(2\lambda \tilde n^{-1}\max_{t\in[p]}|\hat{t}_t(Z_i') - \hat{t}_t(Z_i)|\Big) \le \exp(4\lambda \tilde n^{-1}M). \label{eq:14}
\end{align}

By the mean value theorem, for some $\xi \in (0,4\lambda \tilde n^{-1}M)$,
\begin{equation}
\frac{\hat\omega_{r,s}^{(i)}}{\hat\omega_{r,s}} - 1 \le 4\lambda \tilde n^{-1}M\exp(\xi).
\end{equation}
If $4\lambda \tilde n^{-1}M\le 1$, this gives
\begin{equation}
\frac{\hat\omega_{r,s}^{(i)}}{\hat\omega_{r,s}} - 1 \le 4e\lambda \tilde n^{-1}M,
\end{equation}
and similarly $\frac{\hat\omega_{r,s}^{(i)}}{\hat\omega_{r,s}} - 1 \ge -4\lambda \tilde n^{-1}M$. Thus
\begin{equation}
\Bigg|\frac{\hat\omega_{r,s}^{(i)}}{\hat\omega_{r,s}} - 1\Bigg| \;\le\; 4e\lambda \tilde n^{-1}M.
\end{equation}

Therefore,
\begin{align}
&\Big|Q_{j,r}^{(M),(i)} - \E[Q_{j,r}^{(M),(i)} \mid Z^{(-j),(i)}] - Q_{j,r}^{(M)} + \E[Q_{j,r}^{(M)} \mid Z^{(-j)}]\Big| \notag\\
&\quad\le 4e\lambda \tilde n^{-1}M \sum_{s\ne r}\hat\omega_{r,s}\,\big|\hat{t}_s(Z_j) - \E[\hat{t}_s(Z_j)\mid Z^{(B)}]\big|.
\end{align}

Applying Jensen's inequality,
\begin{align}
&\E\!\Big( Q_{j,r}^{(M),(i)} - \E[Q_{j,r}^{(M),(i)}\mid Z^{(-j),(i)}] - Q_{j,r}^{(M)} + \E[Q_{j,r}^{(M)}\mid Z^{(-j)}]\Big)^2 \notag\\
&\le 16e^2\lambda^2\tilde n^{-2}M^2 \,\E\!\Bigg( \sum_{s\ne r}\hat\omega_{r,s}\,\big(\hat{t}_s(Z_j)-\E[\hat{t}_s(Z_j)\mid Z^{(B)}]\big) \Bigg)^2 \notag\\
&\le 16e^2\lambda^2\tilde n^{-2}M^2 \sum_{s\ne r}\E[\hat\omega_{r,s}]\,\Var(\hat{t}_s(Z_j)) \notag\\
&\le 16e^2\lambda^2\tilde n^{-2}M^2 v^2.
\end{align}

This bound is uniform in $i,j,r$, establishing the first part of \eqref{eq-first-stability}.

\noindent \textbf{Case 2: $i$ and $j$ in different outer folds.}
Here both the weights and nuisance predictions may change. We decompose
\begin{align}
\nabla_i K_{j,r}
&= \sum_{s\neq r} \big(\hat\omega_{r,s}-\hat\omega_{r,s}^{(i)}\big)\,
\big(\hat t_s(Z_j)-\E[\hat t_s(Z_j)\mid Z^{(B)}]\big) \tag{A}\\
&\quad+ \sum_{s\neq r}\hat\omega_{r,s}^{(i)}\,
\Big[(\hat t_s(Z_j)-\E[\hat t_s(Z_j)\mid Z^{(B)}])
-(\hat t_s^{(i)}(Z_j)-\E[\hat t_s^{(i)}(Z_j)\mid Z^{(B),(i)}])\Big]. \tag{B}
\end{align}
For Term (A), we conduct the same procedure in Case 1 to find the ratio bound \eqref{eq:14} becoming
\begin{align}
\frac{\hat \omega_{r,s}^{(i)}}{\hat \omega_{r, s}}
&= \frac{\exp(\lambda \hat\delta_s^{(i)})}{\sum_{t\ne r}\exp(\lambda \hat\delta_t^{(i)})} \cdot \frac{\sum_{t\ne r}\exp(\lambda \hat\delta_t)}{\exp(\lambda \hat\delta_s)} \notag\\
&= \exp(\lambda (\hat \delta_s^{(i)} - \hat \delta_s)) \cdot \frac{\sum_{t \neq r} \exp (\lambda \hat \delta_t^{(i)}) \exp(\lambda (\hat \delta_t - \hat \delta_t^{(i)}))}{\sum_{t \neq r} \exp(\lambda \hat \delta_t^{(i)})} \notag\\
&\le \exp(\lambda(\hat \delta_s^{(i)} - \hat \delta_s)) \cdot \max_t \exp(\lambda (\hat \delta_t - \hat \delta_t^{(i)})) \notag\\
&\le \exp \left(2 \lambda \max_{t \in [p]} \|\hat \delta_t - \hat \delta_t^{(i)}\| \right) \le \exp(2 \lambda \tilde n^{-1} \tilde M),
\end{align}
and the rest is the same as the former case, giving $\|\text{(A)}\|_2 \le C\lambda M^2/n$.  

For Term (B), Assumption~\ref{a4} implies
\[
\|\hat t_s(Z_j)-\E[\hat t_s(Z_j)\mid Z^{(B)}]
-(\hat t_s^{(i)}(Z_j)-\E[\hat t_s^{(i)}(Z_j)\mid Z^{(B),(i)}])\|_2 = o(n^{-1/2}),
\]
uniformly over $s$. Since $\sum_{s\ne r}\hat\omega_{r,s}^{(i)}=1$, this shows $\|\text{(B)}\|_2=o(n^{-1/2})$.  

Combining (A) and (B) completes the proof.
\end{proof}
\section{Proof of Theorem~\ref{thm3}} \label{app-pf2}
\begin{proof} 
To simplify notation, we omit 
the superscript $(-v)$ for every exponential weighting $\hat \omega$ and sample mean $\hat{\delta}$. We also take $r=1$ and define $\tilde{n} = n(1 - 1/V)$. The bounds that we will establish are uniform over $i,j,k,r$. 
    
\textbf{Case 1: \(i\) and \(k\) are in the same outer fold as \(j\), but different inner folds.} The proof of this case mirrors the argument presented in Zhang et al.~\citep{zhang2024winners}. By the definition of $\nabla_i \nabla_k K_{j,1}$ with $i,k \notin I_{v_j}$, one has
\begin{equation}\label{eq: simplify second order}
\begin{aligned}
    &\Enb(\nabla_{i}\nabla_{k} K_{j,1})^2 \\
    &= \Enb \left ( K_{j,1}(\bm{Z}) - K_{j,1}(\bm{Z}^{(k)}) -K_{j,1}(\bm{Z}^{(i)}) + K_{j,1}(\bm{Z}^{(i,k)}) \right )^2 \\
    &= \Enb \Bigg (Q_{j,1} - \E [{Q_{j,1} \mid \bm{Z}^{(-v)}}] - Q_{j,1}^k + \E [{Q_{j,1}^k \mid \bm{Z}^{(-v),(k)}}]  \\
    &\hspace{1cm}+ Q_{j,1}^{i} - \E [{Q_{j,1}^{i} \mid \bm{Z}^{(-v),(i)}}] + Q_{j,1}^{ik} - \E [{Q_{j,1}^{ik} \mid \bm{Z}^{(-v), ({ik})}]} \Bigg )^2 \\
    &= \Enb \left (\sum_{s=2}^p (\hat \omega_{1,s} -\hat \omega_{1,s}^k - \hat \omega_{1,s}^{i} + \hat \omega_{1,s}^{ik}) (\hat{t}_s(Z_j) - \E[\hat{t}_s(Z_j) \mid Z^{(B)}]) \right )^2 .
\end{aligned}
\end{equation}

Consider
\begin{equation}\label{eq: 22}
\begin{aligned}
    &\abs{ \sum_{s = 2}^p (\hat \omega_{1,s}^k - \hat \omega_{1,s} + \hat \omega_{1,s}^{i} - \hat \omega_{1,s}^{ik}) (\hat{t}_s(Z_j) - \E[\hat{t}_s(Z_j) \mid Z^{(B)}]) } \\
    &\le \sum_{s = 2}^p \abs{\hat \omega_{1,s} - \hat \omega_{1,s}^{k} - \hat \omega_{1,s}^{i} + \hat \omega_{1,s}^{ik}} \cdot \abs{\hat{t}_s(Z_j) - \E[\hat{t}_s(Z_j) \mid Z^{(B)}]} \\
    &= \sum_{s = 2}^p \abs{\left(\hat \omega_{1,s}-\hat \omega_{1,s}^{i}\right)\left(1-\frac{\hat \omega_{1,s}^{k}}{\hat \omega_{1,s}}\right)+\hat \omega_{1,s}^{i}\left(\frac{\hat \omega_{1,s}^{ik}}{\hat \omega_{1,s}^{i}}-\frac{\hat \omega_{1,s}^{k}}{\hat \omega_{1,s}}\right)} \\
    &\qquad \cdot \abs{\hat{t}_s(Z_j) - \E[\hat{t}_s(Z_j) \mid Z^{(B)}]} .
\end{aligned}
\end{equation}

The first summation in~\eqref{eq: 22} has $L_2$ norm bounded by $\lambda^2 v M^2 \tilde n^{-2}$.

For the second summation in~\eqref{eq: 22}, consider 
\[
\abs{\hat \omega_{1,s}^{ik}/\hat \omega_{1,s}^{i}-\hat \omega_{1,s}^{k}/\hat \omega_{1,s}} .
\]
Since
\begin{align} \label{shark1}
\frac{\exp (\lambda \hat{\delta}_s^{ik})}{\exp (\lambda \hat{\delta}_s^i)} 
= \frac{\exp (\lambda \hat{\delta}_s^{k})}{\exp (\lambda \hat{\delta}_s)} 
= \exp\!\left(\lambda  \tilde n^{-1}(\hat t_s(Z_k^\prime) - \hat t_s(Z_k))\right),
\end{align}
we obtain
\begin{equation}\label{eq: weight}
\begin{aligned}
    \abs{\frac{\hat \omega_{1,s}^{ik}}{\hat \omega_{1,s}^{i}}-\frac{\hat \omega_{1,s}^{k}}{\hat \omega_{1,s}}}
    &= \exp \left(\lambda \tilde n^{-1}\left(\hat t_s(Z_k^\prime) - \hat t_s(Z_k)\right)\right) 
    \abs{\frac{\Omega}{\Omega^k}\left(\frac{\Omega^i\Omega^k}{\Omega^{ik}\Omega}-1\right)} .
\end{aligned}
\end{equation}
where \(\Omega\) denotes the normalization constant for the exponential weight \(\hat \omega_{1,s}\).
Let
\[
E_{t,t}^{i,k} = \exp(\lambda (\hat{\delta}_t^i + \hat{\delta}_t^k)), \quad 
E_{t,t'}^{i,k} = \exp(\lambda (\hat{\delta}_t^i + \hat{\delta}_{t'}^k)), \quad 
E_{t',t}^{i,k} = \exp(\lambda (\hat{\delta}_{t'}^i + \hat{\delta}_t^k)),
\]
and similarly define $E_{t,t}^{ik,\emptyset}$, $E_{t,t'}^{ik,\emptyset}$ and $E_{t',t}^{ik,\emptyset}$. As $E_{t,t}^{i,k}=E_{t,t}^{ik, \emptyset}$,
\begin{equation}\label{eq: 25}
\begin{aligned}
    \frac{\Omega^i \Omega^k}{\Omega^{ik} \Omega} - 1
    &= 
    \frac{\sum_{t=2}^p E_{t,t}^{i,k} + \sum_{2 \le t < t'} (E_{t,t'}^{i,k} + E_{t',t}^{i,k})}
    {\sum_{t=2}^p E_{t,t}^{ik,\emptyset} + \sum_{2 \le t < t'} (E_{t,t'}^{ik,\emptyset} + E_{t',t}^{ik,\emptyset})}
     - 1 \\
    &\le 
    \sup_{2\le t \le t' \le p} 
    \frac{E_{t,t'}^{i, k}+E_{t',t}^{i,k} - E_{t,t'}^{ik, \emptyset}-E_{t',t}^{ik, \emptyset}}
    {E_{t,t'}^{ik,\emptyset}+E_{t',t}^{ik,\emptyset}} .
\end{aligned}
\end{equation}

For arbitrary $t,t'\in[p]$, the mean value theorem yields
\[
E_{t,t'}^{i, k} - E_{t,t'}^{ik, \emptyset}
= -\exp(\xi_1)\lambda(\hat{\delta}_t^{ik}+\hat{\delta}_{t'}-\hat{\delta}_t^i-\hat{\delta}_{t'}^k),
\]
and similarly
\[
E_{t',t}^{i,k}-E_{t',t}^{ik, \emptyset}
= -\exp(\xi_2)\lambda(\hat{\delta}_{t'}^{ik}+\hat{\delta}_{t}-\hat{\delta}_{t'}^i-\hat{\delta}_t^k).
\]

Thus,
\begin{equation} \label{eq:26}
\begin{aligned}
   &E_{t,t'}^{i, k}+E_{t',t}^{i,k}-E_{t,t'}^{ik, \emptyset}-E_{t',t}^{ik, \emptyset} \\
   &= -\{\exp(\xi_2)-\exp(\xi_1)\}\lambda(\hat{\delta}_{t'}^{ik}+\hat{\delta}_t-\hat{\delta}_{t'}^i-\hat{\delta}_t^k).
\end{aligned}
\end{equation}

The quantity inside parentheses is bounded by $4\lambda \tilde n^{-1}M$. By another application of the mean value theorem, $\exp(\xi_2)-\exp(\xi_1)=\exp(\xi_3)(\xi_2-\xi_1)$ with $\abs{\xi_2-\xi_1}\le 8\lambda \tilde n^{-1}M$. If $4\lambda \tilde n^{-1}M<1$, then
\[
\frac{\exp(\xi_3)}{E_{t,t'}^{ik,\emptyset}+E_{t',t}^{ik,\emptyset}}
\le e .
\]

Consequently,
\[
\sup_{2\le t \le t'\le p}
\frac{E_{t,t'}^{i,k}+E_{t',t}^{i,k}-E_{t,t'}^{ik,\emptyset}-E_{t',t}^{ik,\emptyset}}
{E_{t,t'}^{ik,\emptyset}+E_{t',t}^{ik,\emptyset}}
\le 
32 e \lambda^2 \tilde n^{-2}M^2,
\]
and similarly the infimum is bounded below by 
\[
-\tilde C \lambda^2 \tilde n^{-2}M^2.
\]

Thus,
\[
\abs{\frac{\Omega^i \Omega^k}{\Omega^{ik} \Omega} - 1}
\le C\lambda^2 \tilde n^{-2} M^2.
\]

Returning to~\eqref{eq: weight},
\[
\abs{\frac{\omega_{r,s}^{ik}}{\omega_{r,s}^{i}}-\frac{\omega_{r,s}^{k}}{\omega_{r,s}}}
\le 
\exp (4\lambda \tilde n^{-1}M)
\cdot 
C\lambda^2 \tilde n^{-2}M^2
\le 
C\lambda^2 \tilde n^{-2}M^2.
\]

Hence, by Jensen's inequality, the second summation in~\eqref{eq: 22} has $L_2$ norm of order $O(\lambda^2 v M^2 \tilde n^{-2})$. Combining the bounds for both summations completes the proof.

\textbf{Case 2: \(i\) and \(k\) are in the opposite outer fold of \(j\).} Imitating the proof of Theorem~\ref{thm2}, we decompose
\begin{align}
    \nabla_i \nabla_k K_{j,1} &= \sum_{s=2}^p (\hat \omega_{1,s} -\hat \omega_{1,s}^k - \hat \omega_{1,s}^{i} + \hat \omega_{1,s}^{ik}) (\hat{t}_s(Z_j) - \E[\hat{t}_s(Z_j) \mid Z^{(B)}]) \tag{C}\\
    +& \sum_{s=2}^p (\hat \omega_{1,s}^k - \hat \omega_{1,s}^{ik}) ((\hat t_s(Z_j) - \E[\hat t_s(Z_j) \mid Z^{(B)}]) - (\hat  t_s^k(Z_j) - \E[\hat t_s^k(Z_j) \mid Z^{(B), k}])) \tag{D}\\
    +& \sum_{s=2}^p (\hat \omega_{1,s}^i - \hat \omega_{1,s}^{ik}) ((\hat t_s(Z_j) - \E[\hat t_s(Z_j) \mid Z^{(B)}]) - (\hat t_s^i(Z_j) - \E[\hat t_s^i(Z_j) \mid Z^{(B), i}])) \tag{E}\\
    +& \sum_{s=2}^p \hat \omega_{1,s}^{ik} ((\hat t_s(Z_j) - \E[\hat t_s(Z_j) \mid Z^{(B)}])- (\hat t_s^k(Z_j) - \E[\hat t_s^k(Z_j) \mid Z^{(B), k}]) \notag \\
    &- (\hat t_s^i(Z_j) - \E[\hat t_s^i(Z_j) \mid Z^{(B), i}]) + (\hat t_s^{ik}(Z_j) - \E[\hat t_s^{ik}(Z_j) \mid Z^{(B), ik}])) \tag{F}
\end{align}
Now, term (D) (E) (F) can be controlled by \(o(n^{-1})\) under Assumption~\ref{a4} and the statements in the proof of Theorem~\ref{thm2}. We only need to deal with term (C).

As \(\hat \delta_t^k + \hat \delta_t^i - \hat \delta_t^{ik} -\hat \delta_t = O(n^{-2}) \), Eq~\eqref{eq: 25} holds as well. We first need to make some slight adjustments to Eq~\eqref{shark1}. It now becomes
\begin{align*}
    \frac{\exp(\lambda \hat \delta_s^{ik})}{\exp(\lambda \hat \delta_s^i)} =& \exp \left(\lambda \cdot \tilde n^{-1} \sum_{j=1}^{\tilde n} (\hat t_s^{ik}(Z_j) - \hat t_s^i (Z_j))  \right)\\
    \frac{\exp(\lambda \hat \delta_s^{k})}{\exp(\lambda \hat \delta_s)} =& \exp \left(\lambda \cdot \tilde n^{-1} \sum_{j=1}^{\tilde n} (\hat t_s^{k}(Z_j) - \hat t_s (Z_j))  \right)
\end{align*}
So that
\begin{align*}
    \frac{\exp(\lambda \hat \delta_s^{ik})}{\exp(\lambda \hat \delta_s^i)} = \frac{\exp(\lambda \hat \delta_s^{k})}{\exp(\lambda \hat \delta_s)} \cdot \exp \left(\lambda \cdot (\hat \delta_s - \hat \delta_s^i -\hat \delta_s^k +\hat \delta_s^{ik})  \right)
\end{align*}
By Taylor expansion, we have
\begin{align*}
    \exp \left(\lambda \cdot (\hat \delta_s - \hat \delta_s^i -\hat \delta_s^k +\hat \delta_s^{ik})  \right) =& \lambda \cdot (\hat \delta_s - \hat \delta_s^i -\hat \delta_s^k +\hat \delta_s^{ik}) + 1 + o_P(\lambda(\hat \delta_s - \hat \delta_s^i -\hat \delta_s^k +\hat \delta_s^{ik}))\\
\end{align*}
Now Eq~\eqref{eq: weight} becomes
\begin{align} \label{eq:new15}
    \abs{\frac{\omega_{r,s}^{ik}}{\omega_{r,s}^{i}}-\frac{\omega_{r,s}^{k}}{\omega_{r,s}}} =& \frac{\exp(\lambda \hat \delta_s^{k})}{\exp(\lambda \hat \delta_s)} \abs{\left(\frac{\Omega^i}{\Omega^{ik}}-\frac{\Omega}{\Omega^k}\right)} + [\lambda (\hat \delta_s - \hat \delta_s^i - \hat \delta_s^k + \hat \delta_s^{ik}) \notag\\
    &+ o_P(\lambda (\hat \delta_s - \hat \delta_s^i - \hat \delta_s^k + \hat \delta_s^{ik}))]  \cdot \frac{\sum_{t \neq r} \exp(\lambda \hat \delta_t^{ik})}{\sum_{t \neq r} \exp(\lambda \hat \delta_t^{i})}
\end{align}
Recall the proof of Case 2 in Theorem~\ref{thm2},
\[
\frac{\sum_{t \neq r} \exp(\lambda \hat \delta_t^{ik})}{\sum_{t \neq r} \exp(\lambda \hat \delta_t^{i})} \leq \max_t \exp(\lambda(\hat \delta_t^{ik} - \hat \delta_t^{i})) \leq \lambda \tilde M \tilde n^{-1}
\]
So the last term of Eq~\eqref{eq:new15} can be controlled by \(\lambda^2 \tilde n^{-2}\), which is definitely \(o_P(n^{-1})\).

The next adjustment to be made lies in Eq~\eqref{eq:26}, which now turns to
\begin{align}
    &E_{t, t'}^{i,k} + E_{t',t}^{i,k} - E_{t,t'}^{ik, \emptyset} - E_{t',t}^{ik,\emptyset} \notag\\
        & = -\exp \left(\xi_1\right) \lambda \left(\hat{\delta}_t^{ik}+\hat{\delta}_{t'}-\hat{\delta}_t^i-\hat{\delta}_{t'}^k\right) - \exp \left(\xi_2\right) \lambda\left(\hat{\delta}_{t'}^{ik}+\hat{\delta}_t-\hat{\delta}_{t'}^i-\hat{\delta}_t^k\right) \notag\\
        & = -\exp \left(\xi_1\right) \lambda \left(\hat{\delta}_t^{ik}+\hat{\delta}_{t'}-\hat{\delta}_t^i-\hat{\delta}_{t'}^k\right) - \exp \left(\xi_1\right) \lambda \left(\hat{\delta}_{t'}^{ik}+\hat{\delta}_t-\hat{\delta}_{t'}^i-\hat{\delta}_t^k\right) \notag\\
        &\quad +\exp \left(\xi_1\right) \lambda \left(\hat{\delta}_{t'}^{ik}+\hat{\delta}_t-\hat{\delta}_{t'}^i-\hat{\delta}_t^k\right)-\exp \left(\xi_2\right) \lambda\left(\hat{\delta}_{t'}^{ik}+\hat{\delta}_t-\hat{\delta}_{t'}^i-\hat{\delta}_t^k\right) \notag\\
        & = -\{\exp(\xi_2) - \exp(\xi_1)\}\lambda \left(\hat{\delta}_{t'}^{ik}+\hat{\delta}_t-\hat{\delta}_{t'}^i-\hat{\delta}_t^k\right) - \exp(\xi_1) \lambda [(\hat \delta_{t'}^{ik} + \hat \delta_{t'} - \hat \delta_{t'}^k - \hat \delta_{t'}^i) + (\hat \delta_t + \hat \delta_t^{ik} - \hat \delta_t^i - \hat \delta_t^k)] \tag{G}
\end{align}
Now, let's focus on the last term of (G), we have
\begin{align*}
    &\exp(\xi_1)\lambda [(\hat \delta_{t'}^{ik} + \hat \delta_{t'} - \hat \delta_{t'}^k - \hat \delta_{t'}^i) + (\hat \delta_t + \hat \delta_t^{ik} - \hat \delta_t^i - \hat \delta_t^k)] \leq \exp(\max \{\lambda(\hat \delta_{t}^i + \hat \delta_{t'}^k), \lambda(\hat \delta_{t}^{ik}+\hat \delta_t)\}) \\
    &\cdot \lambda [(\hat \delta_{t'}^{ik} + \hat \delta_{t'} - \hat \delta_{t'}^k - \hat \delta_{t'}^i) + (\hat \delta_t + \hat \delta_t^{ik} - \hat \delta_t^i - \hat \delta_t^k)]\\
\end{align*}
As
\[
\frac{\exp(\max \{\lambda(\hat \delta_{t}^i + \hat \delta_{t'}^k), \lambda(\hat \delta_{t}^{ik}+\hat \delta_t)\})}{E_{t,t'}^{ik, \emptyset}+E_{t',t}^{ik, \emptyset}} \leq e
\]
which is proved in Case 1. The last term of (G) (with the denominator) now can be controlled by \(e \lambda \tilde n^{-2}\), smaller than the first term of (G). The rest of the proof is just analogous to Case 1.

\textbf{Case 3: \(k\) is in the opposite outer fold of \(j\), while \(i\) is in the same outer fold but different inner fold with \(j\).} We can decompose the target into
\begin{align}
    \nabla_i \nabla_k K_{j,1} &= \sum_{s=2}^p (\hat \omega_{1,s} -\hat \omega_{1,s}^k - \hat \omega_{1,s}^{i} + \hat \omega_{1,s}^{ik}) (\hat{t}_s(Z_j) - \E[\hat{t}_s(Z_j) \mid Z^{(B)}]) \tag{H}\\
    +& \sum_{s=2}^p (\hat \omega_{1,s}^k - \hat \omega_{1,s}^{ik}) ((\hat t_s(Z_j) - \E[\hat t_s(Z_j) \mid Z^{(B)}]) - (\hat t_s^k(Z_j) - \E[\hat t_s^k(Z_j) \mid Z^{(B), k}])) \tag{I}
\end{align}
Following the proof in Case 1 and Case 2, term (H) can be controlled by \(\tilde C \lambda^2 \tilde n^{-2}\), where \(\tilde C\) is some constant \(>0\). Imitating the proof of Theorem~\ref{thm2}'s Case 2, term (I) has the upper bound \(\tilde C^\prime \lambda \tilde n^{-2}\). Combining the above gives the proof in Case 3.
\end{proof}

\section{Justification for Assumption~\ref{a4}} \label{appendix:local-stability}
In this section, we demonstrate that a wide range of estimators satisfy Assumption~\ref{a4} with rigorous theoretical guarantees, given local smoothness. The result is summarized as the theorem below.
\begin{theorem}[First- and second-order stability under local smoothness]
\label{thm:local-stability}
Let $S=\{z_i\}_{i=1}^n$ be a training set and let $\hat f_S \in \arg\min_{f\in\mathcal F}\hat L_S(f)$, where
\[
\hat L_S(f) := \frac{1}{n}\sum_{i=1}^n \ell(f;z_i).
\]
For $r\in\{1,\dots,n\}$, let $S^{(r)}$ be obtained by replacing $z_r$ with $z_r'$, and similarly $S^{(r,t)}$ when replacing both $r$ and $t$.
Assume:

\begin{enumerate}
\item \textbf{Per-sample gradient boundedness.} For all $f$ and $z$, $\|\nabla \ell(f;z)\|\le G$.
\item \textbf{Per-sample $\beta$-smoothness.} For all $f,f'$ and $z$, $\|\nabla \ell(f;z)-\nabla \ell(f';z)\|\le \beta\|f-f'\|$ (equivalently $\|\nabla^2\ell(f;z)\|\le \beta$).
\item \textbf{Per-sample Hessian Lipschitzness.} For all $f,f'$ and $z$, $\|\nabla^2 \ell(f;z)-\nabla^2 \ell(f';z)\|\le \rho\|f-f'\|$.
\item \textbf{Local strong convexity of the empirical risk.} There exists a neighborhood $\mathcal N$ of $\hat f_S$ and a constant $\lambda>0$ such that
$\nabla^2 \hat L_S(f)\succeq \lambda I$ for all $f\in\mathcal N$.
\item \textbf{Basin retention.} The ERM solutions $\hat f_{S^{(r)}}$ and $\hat f_{S^{(r,t)}}$ lie in $\mathcal N$ (i.e., they stay in the same local basin where the Hessian is invertible).
\end{enumerate}

Then the ERM is stable in the following sense:
\begin{align}
\|\hat f_{S^{(r)}} - \hat f_S\| &= O(n^{-1}), \label{eq:first}\\
\|\hat f_S - \hat f_{S^{(r)}} - \hat f_{S^{(t)}} + \hat f_{S^{(r,t)}}\| &= O(n^{-2}). \label{eq:second}
\end{align}
The constants implicit in $O(\cdot)$ depend only on $(G,\beta,\rho,\lambda)$ and the basin diameter.
\end{theorem}
\begin{proof}[Proof sketch]
Write $G_S(f):=\nabla \hat L_S(f)$ and $H_S(f):=\nabla^2 \hat L_S(f)$.
First-order optimality gives $G_S(\hat f_S)=0$ and $G_{S^{(r)}}(\hat f_{S^{(r)}})=0$.

\noindent \textbf{Step 1: Single replacement (first-order stability).}
Apply a first-order Taylor expansion of $G_{S^{(r)}}$ at $\hat f_S$:
\[
0 = G_{S^{(r)}}(\hat f_{S^{(r)}}) = G_{S^{(r)}}(\hat f_S) + H_{S^{(r)}}(\Omega)\,(\hat f_{S^{(r)}}-\hat f_S),
\]
for some $\Omega$ on the segment between $\hat f_S$ and $\hat f_{S^{(r)}}$.
Add and subtract $G_S(\hat f_S)=0$ and rewrite the Hessian at the anchor $H_S(\hat f_S)$:
\[
0
= \underbrace{\bigl(G_{S^{(r)}}(\hat f_S)-G_S(\hat f_S)\bigr)}_{:=\ \delta_S^{(r)}}
+ H_S(\hat f_S)(\hat f_{S^{(r)}}-\hat f_S)
+ \underbrace{\bigl(H_{S^{(r)}}(\Omega)-H_S(\hat f_S)\bigr)}_{:=\ \Delta H}\,(\hat f_{S^{(r)}}-\hat f_S).
\]
By (A1), replacing one sample perturbs the empirical gradient by at most
\[
\|\delta_S^{(r)}\| = \left\|\frac{1}{n}\bigl(\nabla \ell(\hat f_S;z_r')-\nabla \ell(\hat f_S;z_r)\bigr)\right\| \le \frac{2G}{n} = O(n^{-1}).
\]
By (A2)–(A3) and linearity of the Hessian in the empirical measure,
\[
\|H_{S^{(r)}}(f)-H_S(f)\|\le \frac{2\beta}{n},\quad
\|H_S(\Omega)-H_S(\hat f_S)\|\le \rho\|\hat f_{S^{(r)}}-\hat f_S\|.
\]
Thus
\[
\|\Delta H\| \le \frac{2\beta}{n} + \rho\|\hat f_{S^{(r)}}-\hat f_S\|.
\]
Let $\Delta_r:=\hat f_{S^{(r)}}-\hat f_S$. Using (A4), $H_S(\hat f_S)$ is invertible and
$\|H_S(\hat f_S)^{-1}\|\le \lambda^{-1}$. Solve for $\Delta_r$:
\[
\Delta_r = -H_S(\hat f_S)^{-1}\delta_S^{(r)} - H_S(\hat f_S)^{-1}\Delta H\,\Delta_r.
\]
Taking norms and rearranging,
\[
\|\Delta_r\| \le \frac{1}{\lambda}\|\delta_S^{(r)}\|
+ \frac{1}{\lambda}\|\Delta H\|\,\|\Delta_r\|
\le \frac{C_1}{n} + \frac{1}{\lambda}\Big(\frac{2\beta}{n}+\rho\|\Delta_r\|\Big)\|\Delta_r\|.
\]
For $n$ large enough (so that $\|\Delta_r\|$ is small within the basin), the quadratic term can be absorbed, yielding
$\|\Delta_r\| \le C/n$, proving \eqref{eq:first}.

\noindent \textbf{Step 2: Double replacement (second-order stability).}
Repeat Step~1 for $S^{(t)}$ and $S^{(r,t)}$ to obtain
\[
\hat f_{S^{(r)}}-\hat f_S = -H_S^{-1}\delta_S^{(r)} + R_S^{(r)},\quad
\hat f_{S^{(t)}}-\hat f_S = -H_S^{-1}\delta_S^{(t)} + R_S^{(t)},\quad
\hat f_{S^{(r,t)}}-\hat f_S = -H_S^{-1}\delta_S^{(r,t)} + R_S^{(r,t)},
\]
where $H_S := H_S(\hat f_S)$ and each remainder $R_S^{(\cdot)}$ is $O(n^{-1}\|\hat f_{(\cdot)}-\hat f_S\|)=O(n^{-2})$ by the bound on $\Delta H$ and Step~1.
Form the finite-difference combination:
\[
\hat f_S - \hat f_{S^{(r)}} - \hat f_{S^{(t)}} + \hat f_{S^{(r,t)}}
= -H_S^{-1}\Big(\delta_S^{(r,t)}-\delta_S^{(r)}-\delta_S^{(t)}\Big)
+ \Big(R_S^{(r,t)}-R_S^{(r)}-R_S^{(t)}\Big).
\]
Since all $\delta$’s are evaluated at the same anchor $\hat f_S$ and the empirical gradient is linear in the sample measure,
\[
\delta_S^{(r,t)}-\delta_S^{(r)}-\delta_S^{(t)} = 0.
\]
Thus the entire term is controlled by the $O(n^{-2})$ remainders, yielding \eqref{eq:second}.
\end{proof}
\section{Experimental Details}\label{appe:sec:experiment}
\subsection{The Toy Model} \label{app-toy}
\noindent \textbf{Simulated Dataset.} For simplicity, we use simulated data with linear outcomes here. We sample latent covariates $L \sim \mathcal{N}(0, I_m)$ and split them into 
$I, C, A, D$. The treatment assignment depends on $IC = (I, C)$ through
\(
e(Z) = \sigma(IC \cdot w_{IC} + \varepsilon), 
\; \varepsilon \sim \mathcal{N}(0, 1),
\)
with clipping to $[0.1, 0.9]$, and
\(
W \sim \mathrm{Bernoulli}(t(Z)).
\)
Potential outcomes follow
\(f_0 = \frac{(C, A)^{\circ 1} w_0}{m_C + m_A} + \eta_0, 
\;
f_1 = \frac{(C, A)^{\circ 1} w_1}{m_C + m_A} + \eta_1,
\)
where $\eta_0, \eta_1 \sim \mathcal{N}(0, 0.5)$. We generate  $\mu_0 = f_0$ and $\mu_1 = f_1$.

\noindent \textbf{Nuisance Estimators.} We use linear regression to predict conditional outcomes and logistic regression to predict the propensity score.

\noindent \textbf{Candidate Estimators.} Denote the real treatment effect as \(\tau(X)\). The candidate estimators are generated by perturbing the true treatment effect $\tau$ with independent Gaussian noise, i.e. \(\hat \tau_i(X) = \tau(X) + \epsilon_i\). We use different settings in Figure~\ref{fig:naive_ours} and Figure~\ref{fig:split_difference}.
\begin{itemize}
\item For the former one, \(\epsilon_1 \sim \mathcal{N} (0, 0.1), \; \epsilon_2 \sim \mathcal{N} (0.03, 0.1), \; \epsilon_3 \sim \mathcal{N} (0.03, 0.1), \; \epsilon_4 \sim \mathcal{N} (0.3, 0.1) ,\;
\epsilon_5 \sim \mathcal{N}(0.3, 0.1),\;
\epsilon_6 \sim \mathcal{N}(0.3, 0.1),\;
\epsilon_7 \sim \mathcal{N}(0.3, 0.1) \) .
\item For the latter one, \(\epsilon_1 \sim \mathcal{N} (0, 0.1), \; \epsilon_2 \sim \mathcal{N} (0.03, 0.1), \; \epsilon_3 \sim \mathcal{N} (0.03, 0.1), \; \epsilon_4 \sim \mathcal{N} (0.03, 0.1) ,\;
\epsilon_5 \sim \mathcal{N}(0.03, 0.1) \) .
\end{itemize}

\subsection{Dataset Details} \label{app-dataset}
\noindent \textbf{ACIC 2016.} The ACIC 2016 competition dataset~\cite{dorie2019automated}, built upon the real-world Collaborative Perinatal Project~\cite{niswander1972women}, contains 55 observed covariates of diverse types with realistic dependence structures, while treatment assignments and potential outcomes are synthetically generated under multiple data-generating mechanisms. We use the R package \texttt{aciccomp2016} with setting~1 to generate semi-synthetic datasets.

\noindent \textbf{IHDP.} The IHDP dataset originates from a randomized experiment evaluating whether specialist home visits improve long-term cognitive outcomes for children. We adopt the semi-synthetic version of Hill~\cite{Hill01012011}, where a subset of treated individuals is removed to induce selection bias, yielding 747 observations (139 treated and 608 control) with 25 pre-treatment features. Outcomes are generated using the ``A'' specification in the NPCI package~\cite{dorie2016npci} following Shalit et al.~\cite{pmlr-v70-shalit17a}.

\noindent \textbf{Twins.} The Twins dataset is derived from U.S.\ twin birth records, labeling the heavier infant as treated ($t_i=1$) and the lighter as control ($t_i=0$). We retain 28 parental, prenatal, and birth-related covariates, augmented with 10 additional features following~\cite{wu2022learning}. The outcome is the one-year mortality indicator. Restricting to same-sex twins with birth weights below 2000g and no missing covariates yields 5,271 samples. Treatment assignment follows
\[
t_i \mid x_i \sim \mathrm{Bern}\!\left(\sigma(w^\top x_i + n)\right),
\]
where $\sigma$ is the logistic function, $w \sim \mathcal{U}((-0.1,0.1)^{38\times 1})$, and $n \sim \mathcal{N}(0,0.1)$.

\subsection{Details of the Candidate Estimators} \label{app-candidate-details}
We choose Causal Forest with different hyperparameters. The values are listed in Table~\ref{tab:estimator-configs}.

\begin{table}[h]
\centering
\caption{Hyperparameter configurations of the seven candidate Causal Forest estimators across datasets.}
\label{tab:estimator-configs}
\begin{tabular}{lcccc}
\toprule
& \texttt{n\_estimators} & \texttt{t\_n\_estimators} & \texttt{y\_n\_estimators} & \texttt{max\_depth} \\
\midrule
\multicolumn{5}{l}{\textit{Twins}} \\
\midrule
Model 1 & 232 & 40 & 40 & 8 \\
Model 2 & 152 & 30 & 30 & 12 \\
Model 3 & 260 & 50 & 50 & 18 \\
Model 4 & 120 & 30 & 30 & 8 \\
Model 5 & 180 & 45 & 45 & 12 \\
Model 6 & 160 & 40 & 40 & 10 \\
Model 7 & 152 & 30 & 30 & 8 \\
\midrule
\multicolumn{5}{l}{\textit{IHDP}} \\
\midrule
Model 1 & 96  & 14 & 14 & 3 \\
Model 2 & 40  & 12 & 12 & 3 \\
Model 3 & 40  & 10 & 10 & 3 \\
Model 4 & 96  & 5  & 5  & 3 \\
Model 5 & 160 & 30 & 30 & 8 \\
Model 6 & 52  & 10 & 10 & 3 \\
Model 7 & 60  & 20 & 20 & 4 \\
\midrule
\multicolumn{5}{l}{\textit{ACIC}} \\
\midrule
Model 1 & 200 & 120 & 120 & 10 \\
Model 2 & 200 & 120 & 120 & 8  \\
Model 3 & 200 & 150 & 150 & 10 \\
Model 4 & 180 & 150 & 150 & 10 \\
Model 5 & 200 & 120 & 150 & 10 \\
Model 6 & 200 & 150 & 150 & 8  \\
Model 7 & 220 & 150 & 150 & 10 \\
\bottomrule
\end{tabular}
\end{table}

\subsection{Experiments with Diverse Candidate Estimators}
\label{app-diverse}

To assess the effectiveness of our method beyond Causal Forest variants, we conduct a supplementary experiment using a heterogeneous set of five candidate estimators: Causal Forest, TARNet, X-learner, S-learner, and LinearDML. This experiment is evaluated on ACIC 2016 and Twins over 50 independent replications, reporting the same metrics as Table~\ref{tab:fwer}. While our method consistently outperforms existing baselines, the performance gap narrows on Twins, where candidates display larger performance differences and any sensible selection method performs reasonably well.

\begin{table}[h]
\centering
\caption{FWER and average number of false selections (ANFS) with diverse candidate estimators (mean $\pm$ standard error over 50 repetitions).}
\label{tab:diverse}
\begin{tabular}{lcccc}
\toprule
& \multicolumn{2}{c}{\textit{ACIC}} & \multicolumn{2}{c}{\textit{Twins}} \\
\cmidrule(lr){2-3} \cmidrule(lr){4-5}
Method & FWER & ANFS & FWER & ANFS \\
\midrule
Ours             & 0 & $0.12 \pm 0.05$ & 0 & $0.04 \pm 0.03$ \\
Naive            & 0 & $0.24 \pm 0.06$ & 0 & $0.06 \pm 0.03$ \\
Ranking Inference & 0 & $0.80 \pm 0.06$ & 0 & $0.12 \pm 0.05$ \\
Bonferroni       & 0 & $0.72 \pm 0.06$ & 0 & $0.08 \pm 0.04$ \\
\bottomrule
\end{tabular}
\end{table}

\section{Nuisance Estimator Reliability} \label{app-nuisance}
\begin{figure}[h]
\centering
\begin{minipage}[t]{0.45\textwidth}
    \centering
    \includegraphics[width=\textwidth]{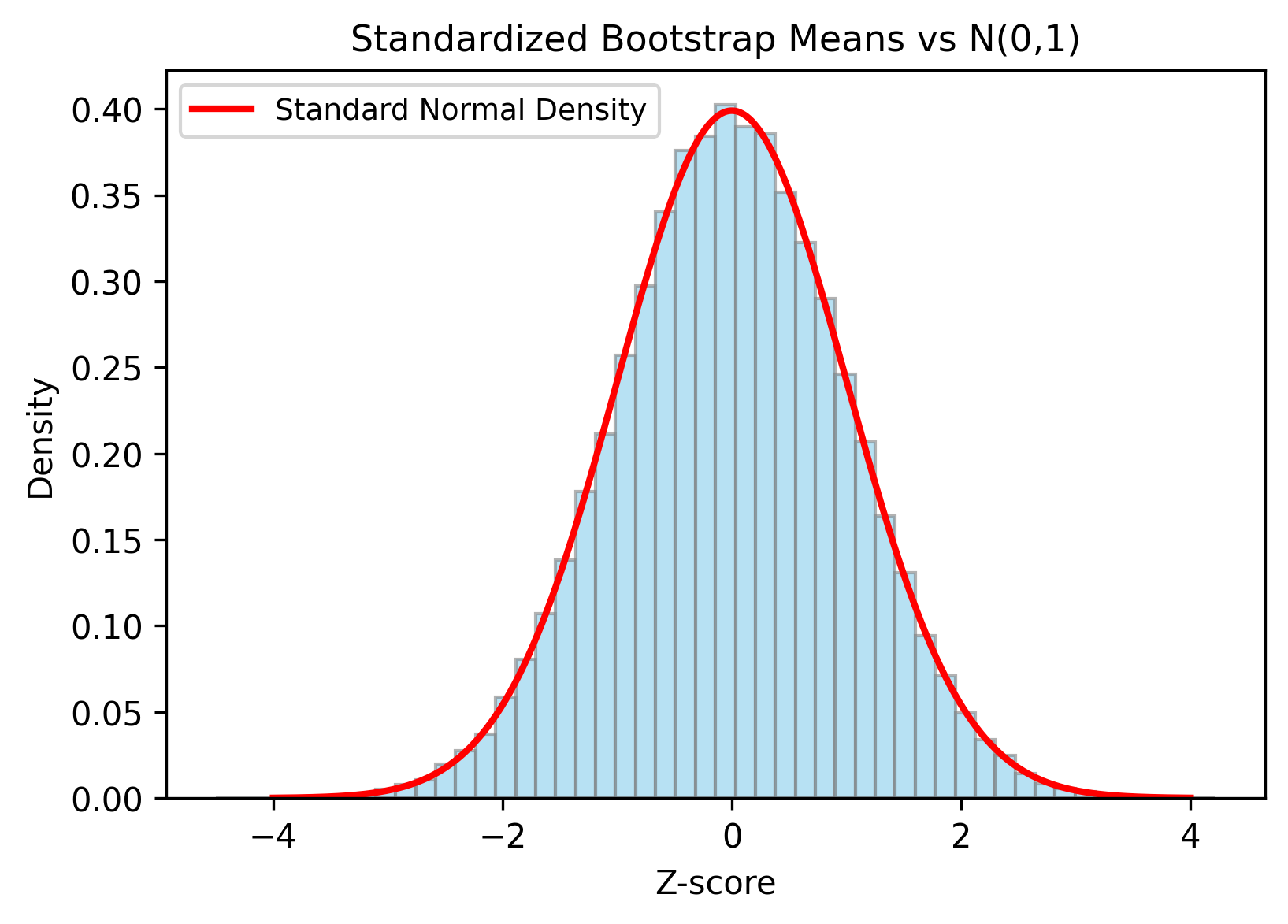}
    \vspace{-4pt}
    \caption*{(a) Bootstrap mean distribution of one pair vs $N(0,1)$.}
\end{minipage}\hfill
\begin{minipage}[t]{0.45\textwidth}
    \centering
    \includegraphics[width=\textwidth]{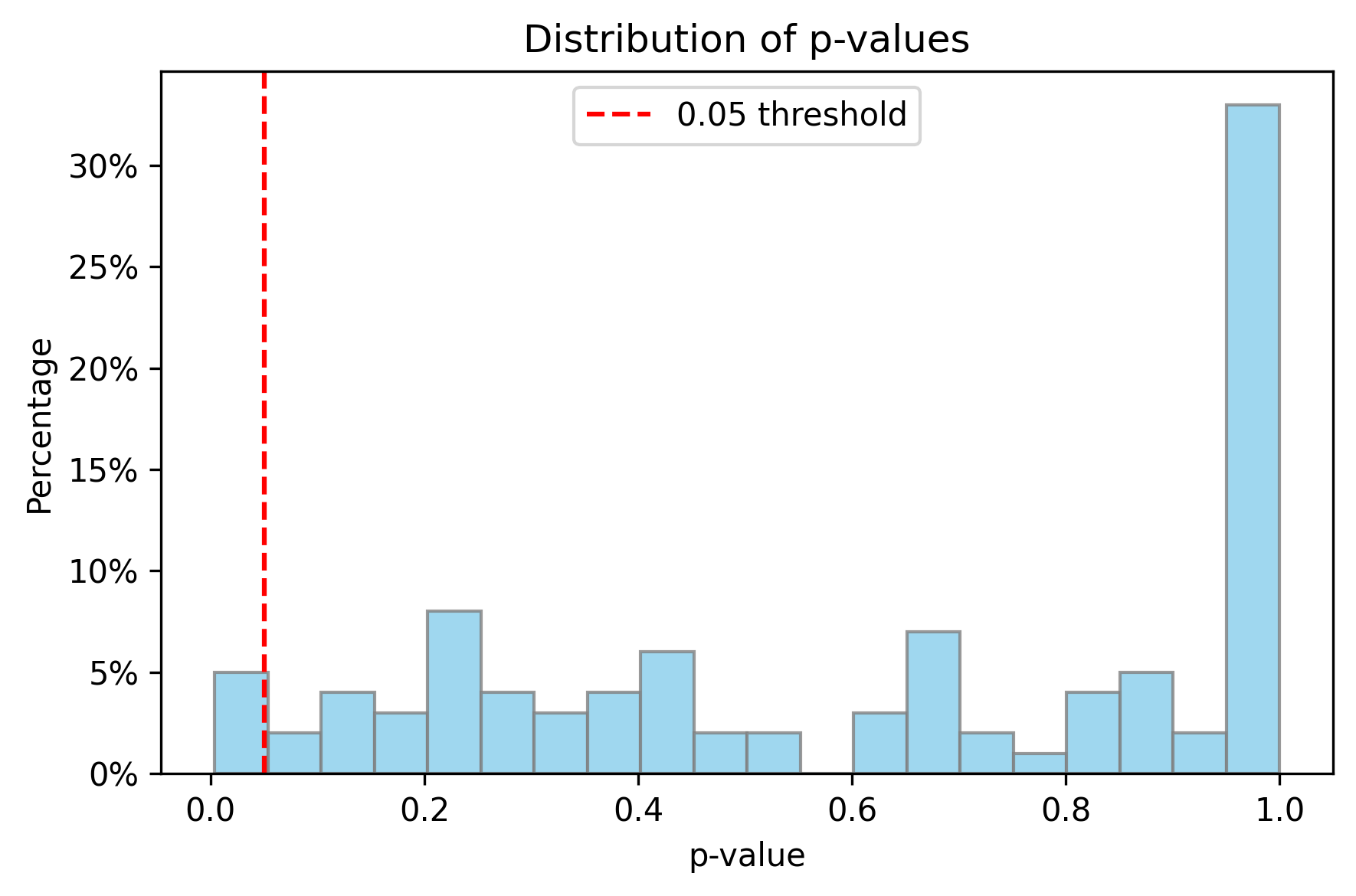}
    \vspace{-4pt}
    \caption*{(b) Bonferroni-adjusted $p$-values across 100 datasets.}
\end{minipage}
\vspace{4pt}
\caption{Illustration of the CLT validity diagnostics. 
(a) shows the bootstrap mean distribution for one representative pair compared to the standard normal. 
(b) summarizes Bonferroni-adjusted $p$-values of KS test for all pairs across 100 datasets, indicating no evidence against the CLT approximation.}
\label{fig:clt-visualization}
\vspace{-8pt}
\end{figure}

We empirically demonstrate that the nuisance estimators we adopt indeed
lead to test statistics that are approximately Gaussian.  
We present results on the ACIC 2016 datasets.  
In Figure~\ref{fig:clt-visualization}, panel (a) shows a bootstrap approximation of the sampling distribution
for one representative pair of estimators; the resulting histogram aligns
closely with the shape of a standard normal distribution.  
Panel (b) further summarizes the behavior across all pairs: for each of
the 100 datasets, we perform a KS test for normality and report the
Bonferroni-corrected $p$-values. As shown in the bar plot, approximately
5\% of the datasets have corrected $p$-values below the 0.05 threshold,
which is exactly what one would expect under the null.  
Together, these results confirm that our test statistics satisfy the asymptotic normality predicted by the
CLT.

\end{document}